\documentclass[10pt,twocolumn,letterpaper]{article}

\usepackage{iccv}
\usepackage{times}
\usepackage{epsfig}
\usepackage{graphicx}
\usepackage{amsmath}
\usepackage{amssymb}
\usepackage{enumerate}
\usepackage{multirow}
\usepackage{booktabs}
\usepackage{bbding}
\usepackage{colortbl}
\usepackage{xcolor}
\usepackage{array}
\newcommand{\subsubsubsection}[1]{\paragraph{#1}\mbox{}\\}
\newcommand{\tabincell}[2]{\begin{tabular}{@{}#1@{}}#2\end{tabular}} 
\usepackage[pagebackref=true,breaklinks=true,letterpaper=true,colorlinks,bookmarks=false]{hyperref}
\iccvfinalcopy 

\def\iccvPaperID{2748} 

\begin{document}

\title{DC-Net: Divide-and-Conquer for Salient Object Detection}

\author{Jiayi Zhu\\
MBZUAI\\
Abu Dhabi, UAE\\
{\tt\small zjyzhujiayi55@gmail.com}
\and
Xuebin Qin\\
MBZUAI\\
Abu Dhabi, UAE\\
{\tt\small xuebinua@gmail.com}
\and
Abdulmotaleb Elsaddik\\
MBZUAI\\
Abu Dhabi, UAE\\
{\tt\small a.elsaddik@mbzuai.ac.ae}
}

\maketitle
\ificcvfinal\thispagestyle{empty}\fi

\begin{abstract}
In this paper, to guide the model's training process to explicitly present a progressive trend, we first introduce the concept of Divide-and-Conquer into Salient Object Detection (SOD) tasks, called DC-Net. Our DC-Net guides multiple encoders to solve different subtasks and then aggregates the feature maps with different semantic information obtained by multiple encoders into the decoder to predict the final saliency map. The decoder of DC-Net consists of newly designed two-level Residual nested-ASPP (ResASPP$^{2}$) modules, which improve the sparse receptive field existing in ASPP and the disadvantage that the U-shape structure needs downsampling to obtain a large receptive field. Based on the advantage of Divide-and-Conquer's parallel computing, we parallelize DC-Net through reparameterization, achieving competitive performance on six LR-SOD and five HR-SOD datasets under high efficiency (60 FPS and 55 FPS). Codes and results are available: \href{https://github.com/PiggyJerry/DC-Net}{https://github.com/PiggyJerry/DC-Net}.
\end{abstract}

\section{Introduction}
Recently, with the development of deep convolutional neural networks (CNNs), downstream computer vision tasks have been greatly improved, and Salient Object Detection (SOD) has also benefited from it. The purpose of SOD is to segment the most visually attractive part of an image, and it is widely used in 3D modeling, image editing, art design materials, 
AR and 3D rendering. So what are the deficiencies worthy of researchers to explore? Next, we will discuss it based on the previous method.

\begin{figure}[t]
\centering
\includegraphics[scale=.5]{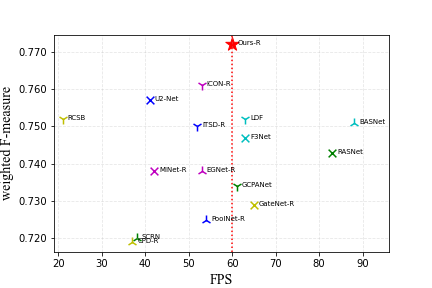}
\caption{Comparison of FPS and performance of our DC-Net-R with other state-of-the-art SOD convolution-based methods. The $F_{\beta}^{w}$ measure is computed on dataset DUT-OMRON \cite{yang2013saliency}. The red star denotes our DC-Net-R (Ours-R, 60 FPS) and the red dot line denotes the real-time (60 FPS) line.}
\label{fig:highlight}
\vspace{-0.5em}
\end{figure}

In recent years, several deep salient object detection methods have introduced different auxiliary maps (e.g. edge maps, body maps, and detail maps) to assist in generating saliency maps, and their designs fall into the following three categories. First, after feeding the image into an encoder, use the features learned by predicting different auxiliary maps to assist in predicting the saliency maps \cite{liu2019simple,zhao2019egnet}.
The second is to use auxiliary maps as input to guide the training process \cite{tu2020edge}. The third is to make the models pay more attention to the edge pixels through the boundary-aware loss \cite{qin2019basnet,feng2019attentive}. However, these methods have some limitations. For the first method, a single encoder with multiple heads to learn different semantic information may not fully represent all the different semantic information \cite{wei2020label, wu2019stacked}. Moreover, when multiple branches need to interact with each other with a sequence, they cannot be accelerated through parallelism, leading to low efficiency \cite{zhou2020interactive}. 
The second method suffers from the need to generate auxiliary maps during the inference stage, leading to low efficiency. The third method can only use the boundary information. It is more intuitive and effective to directly use auxiliary maps for training. Moreover, most methods directly input the output feature maps of a single encoder into the decoder to predict the final saliency map. However, the output feature maps of a single encoder are a fusion of different features and cannot fully consider the quality of each feature.

\begin{figure*}[t]
\centering
\includegraphics[width=\textwidth]{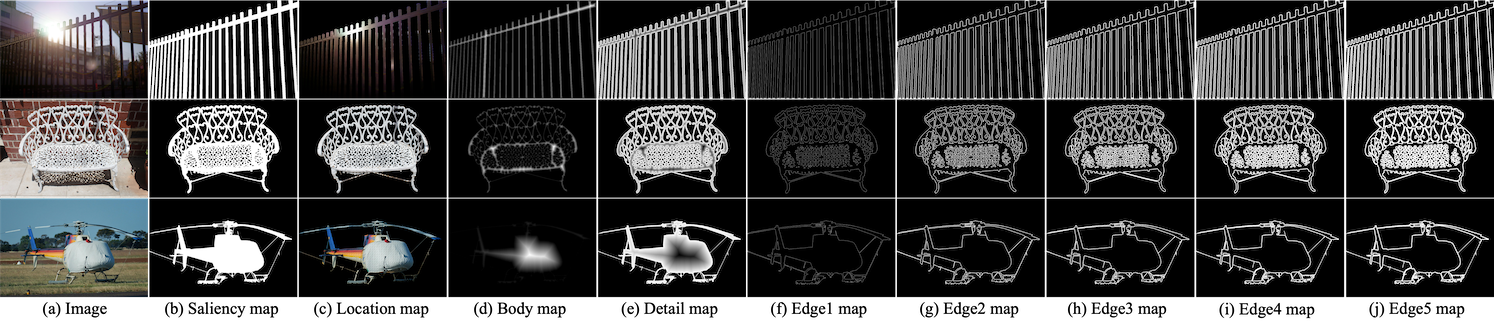}
\caption{Some examples of different auxiliary maps. (c) represents the location information of the salient object. The sum of (d) and (e) is equal to (b). (f)-(j) represents the edge pixels of salient objects with widths 1, 2, 3, 4, and 5 respectively.}
\label{fig:auxiliary maps}
\vspace{-0.5em}
\end{figure*}

This leads to our first question: \textbf{can we design an end-to-end network that explicitly guides the training process by using multiple encoders to represent the different semantic information of the saliency map to learn the prior knowledge before predicting the saliency map and be efficient?}

Current mainstream pixel-level deep CNNs such as U-Net \cite{ronneberger2015u} and feature pyramid network (FPN) \cite{lin2017feature} increase the receptive field and improve efficiency through continuous pooling layers or convolutional layers with stride of 2, while pooling operation can lose detail information, that is, sacrifice the high resolution of the feature maps, and convolution with stride of 2 results in no convolution operation on half the pixels. Dilated convolution and atrous spatial pyramid pooling (ASPP) are proposed by Deeplab \cite{chen2017deeplab} for this problem, but due to the large gap in the atrous rate and only one parallel convolutional layer, the pixel sampling is sparse. A recently proposed method U$^{2}$-Net \cite{qin2020u2} proposes a ReSidual U-blocks (RSU), and it can obtain multi-scale feature maps after several pooling layers at each stage, and finally restore to the high resolution of the current stage like U-Net, but the pooling operation still leads to the loss of detail information in this process.

Therefore, our second question is: \textbf{can we design a module to obtain a larger receptive field with fewer convolutional layers while maintaining the high resolution of the feature maps of the current stage all the time?}

Our main contribution is a novel method for SOD, called Divide-and-Conquer Network (\textbf{DC-Net}) with a two-level Residual nested-ASPP module (\textbf{ResASPP$^{2}$}), which solves the two issues raised above, and we introduce \textbf{Parallel Acceleration} into DC-Net to speed it up. Our network training process is as follows: after feeding the image into two identical encoders, edge maps with width 4 and the location maps are used to supervise the two encoders respectively, as shown in Fig. \ref{fig:auxiliary maps} (i) and (c), and then the concatenation of the feature maps of the two encoders are fed into the decoder composed of ResASPP$^{2}$s to predict the final saliency maps in the way of U-Net like structure. ResASPP$^{2}$ obtains a large and compact effective receptive field (ERF) without sacrificing high resolution by nesting two layers of parallel convolutional layers with dilation rates \{1, 3, 5, 7\}. Additionally, its output feature map has much diversity by fusing a large number of feature maps with different scales and compact pixel sampling. Parallel Acceleration merges two identical encoders into an encoder with the same structure
, which is called \textbf{Parallel Encoder}. ResASPP$^{2}$ is simplified by our implementation of \textbf{Merged Convolution}. Our DC-Net achieves competitive performance against the state-of-the-art (SOTA) methods on five public SOD datasets
and runs at real-time (60 FPS based on \textbf{Parallel-ResNet-34}, with input size of $352\times 352\times 3$; 55 FPS based on \textbf{Parallel-ResNet-34}, with input size of $1024\times 1024\times 3$; 29 FPS based on \textbf{Parallel-Swin-B}, with input size of $384\times 384\times 3$) on a single RTX 6000 GPU.

\section{Related Works}
Under the increasing demand for higher efficiency and accuracy in the real world, traditional methods \cite{jiang2013saliency,yang2013saliency,lu2013robust}
based on hand-crafted features are gradually losing competitiveness. In recent years, more and more deep salient object detection networks \cite{hou2017deeply,zhang2017amulet}
have been proposed, and a lot of research has been done on how to integrate multi-level and multi-scale features \cite{li2021cross}, and how to use the auxiliary maps such as the edge map to train the network \cite{zhao2019egnet}. Recently, the emergence of SAM \cite{kirillov2023segment} and its variants, such as MedSAM \cite{ma2023segment} and HQ-SAM \cite{ke2023segment}, has greatly facilitated the development of segmentation tasks. However, research on the aforementioned issues remains crucial for achieving better performance.

\begin{figure*}[t]
\centering
\includegraphics[width=\textwidth]{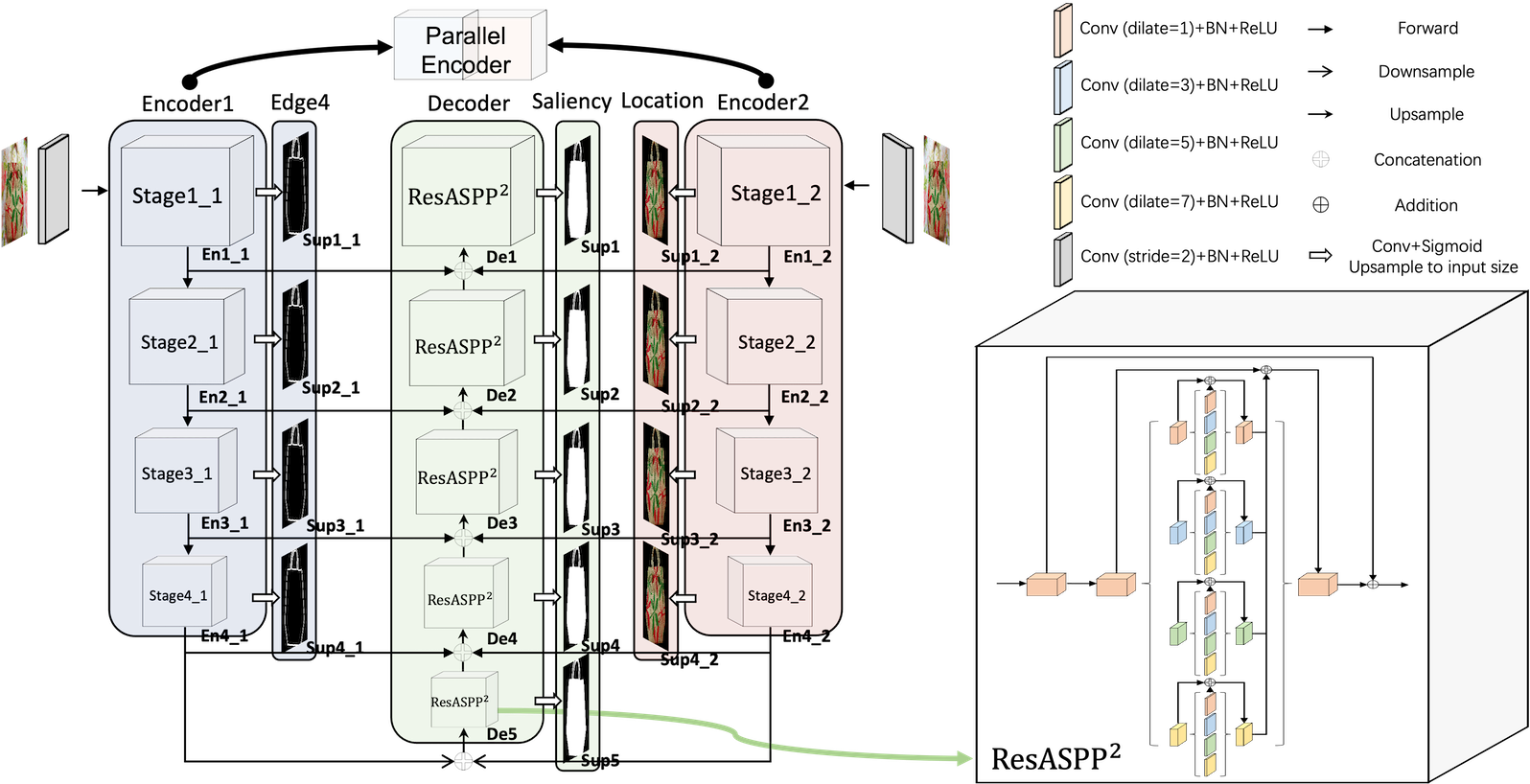}
\caption{Illustration of our proposed DC-Net architecture. DC-Net has two encoders and a decoder, we can consider these two encoders as one parallel encoder. Thus, the main architecture of DC-Net is a U-Net like Encoder-Decoder, where each stage of the decoder consists of our newly proposed two-level Residual nested-ASPP module (ResASPP$^{2}$).}
\label{fig:DC-Net}
\end{figure*}

\textbf{Multi-level and multi-scale feature integration:} 
Recent works such as U-Net \cite{ronneberger2015u}, Feature Pyramid Network (FPN) \cite{lin2017feature}, PSPNet \cite{zhao2017pyramid} and Deeplab \cite{chen2017deeplab} have shown that the fusion of multi-scale contextual features can lead to better results. Many subsequent developed methods for SOD to integrate or aggregate multi-level and multi-scale features were inspired by them to some extent. 
Liu \textit{et al.} (PoolNet) \cite{liu2019simple} aggregate the multi-scale features obtained from a module adapted from pyramid pooling module at each level of the decoder and a global guidance module is introduced to help each level obtain better location information.
Wei \textit{et al.} (F$^{3}$Net) \cite{wei2020f3net} propose a feature fusion strategy that is different from addition or concatenation, which can adaptively select fused features and reduce redundant information.
Mohammadi \textit{et al.} (CAGNet) \cite{mohammadi2020cagnet} propose a multi-scale feature extraction module that combines convolutions with different sizes of convolution kernels in parallel.
Zhao \textit{et al.} (GateNet) \cite{zhao2020suppress} propose a Fold-ASPP module to generate finer multi-scale advanced saliency features.
Zhuge \textit{et al.} (ICON) \cite{zhuge2022salient} make full use of the features under various receptive fields to improve the diversity of features, and introduce an attention module to enhance feature channels that the integral salient objects are highlighted.
Liu \textit{et al.} (PiCANet) \cite{liu2020picanet} generate global and local contextual attention for each pixel and use it on a U-Net structure.
Li \textit{et al.} \cite{li2023robust} design a dynamic searching process module as a meta operation to conduct multi-scale and multi-modal feature fusion.
Liu \textit{et al.} (SMAC) \cite{liu2021learning} propose a novel mutual attention model by fusing attention and context from different modalities to better fuse cross-modal information.
Fang \textit{et al.} (DNTDF) \cite{fang2022densely} propose the progressive compression shortcut paths (PCSPs) to read features from higher levels.
Zhang \textit{et al.} \cite{zhang2020revisiting} propose a multi-scale, multi-modal, and multi-level feature fusion module, leveraging the robustness of the thermal sensing modality to illumination and occlusion.
Chen \textit{et al.} (GCPANet) \cite{chen2020global} integrate global context features with low- and high-level features.
Wu \textit{et al.} (CPD) \cite{wu2019cascaded} propose a framework for fast and accurate salient object detection named Cascaded Partial Decoder.
Pang \textit{et al.} (MINet) \cite{pang2020multi} propose the aggregate interaction modules and self-interaction modules to integrate the features from adjacent levels and obtain more efficient multi-scale features from the integrated features.
Chen \textit{et al.} (RASNet) \cite{chen2020reverse} employ
residual learning to refine saliency maps progressively and design a novel top-down
reverse attention block to guide the residual
learning.
Qin \textit{et al.} (U$^{2}$-Net) \cite{qin2020u2} propose ReSidual U-blocks (RSU) to capture more contextual information from different scales and increase the depth of the whole architecture without significantly increasing the computational cost.
Xie \textit{et al.} (PGNet) \cite{xie2022pyramid} integrates the features extracted by the Transformer and CNN backbones, enabling the network to combine the detection ability of Transformer with the detailed representation ability of CNN.

\textbf{Utilizing auxiliary supervision:} Many auxiliary maps such as edge maps, body maps and detail maps have been introduced to assist in predicting the saliency map for SOD in recent years.
Liu \textit{et al.} (PoolNet) \cite{liu2019simple} fuses edge information with saliency predictions in a multi-task training manner. 
Zhao \textit{et al.} (EGNet) \cite{zhao2019egnet} perform interactive fusion after explicit modeling of salient objects and edges to jointly optimize the tasks of salient object detection and edge detection under the belief that these two tasks are complementary. 
Qin \textit{et al.} (BASNet) \cite{qin2019basnet}propose a hybrid loss which can focus on the pixel-level, patch-level, and map-level salient parts of the image.
Su \textit{et al.} (BANet) \cite{su2019selectivity} use the selective features of boundaries to slight appearance change to distinguish salient objects and background.
Feng \textit{et al.} (AFNet) \cite{feng2019attentive} design the Attentive Feedback Modules (AFMs) and a Boundary-Enhanced Loss (BEL) to better explore the structure of objects and learn exquisite boundaries respectively.
Wu \textit{et al.} (SCRN) \cite{wu2019stacked} propose a stacking Cross Refinement Unit (CRU) to simultaneously refine multi-level features of salient object detection and edge detection.
Wei \textit{et al.} (LDF) \cite{wei2020label} explicitly decompose the original saliency map into body map and detail map so that edge pixels and region pixels have a more balanced distribution.
Ke \textit{et al.} (RCSB) \cite{ke2022recursive} propose a contour-saliency blending module to exchange information between contour and saliency.
Zhou \textit{et al.} (ITSD) \cite{zhou2020interactive} propose an interactive two-stream decoder to explore multiple cues, including saliency, contour and their correlation.
Qin \textit{et al.} (IS-Net) \cite{qin2022highly} propose a simple intermediate supervision baseline using both feature-level and mask-level guidance for model training.

\begin{figure*}[t]
\centering
\includegraphics[width=\textwidth]{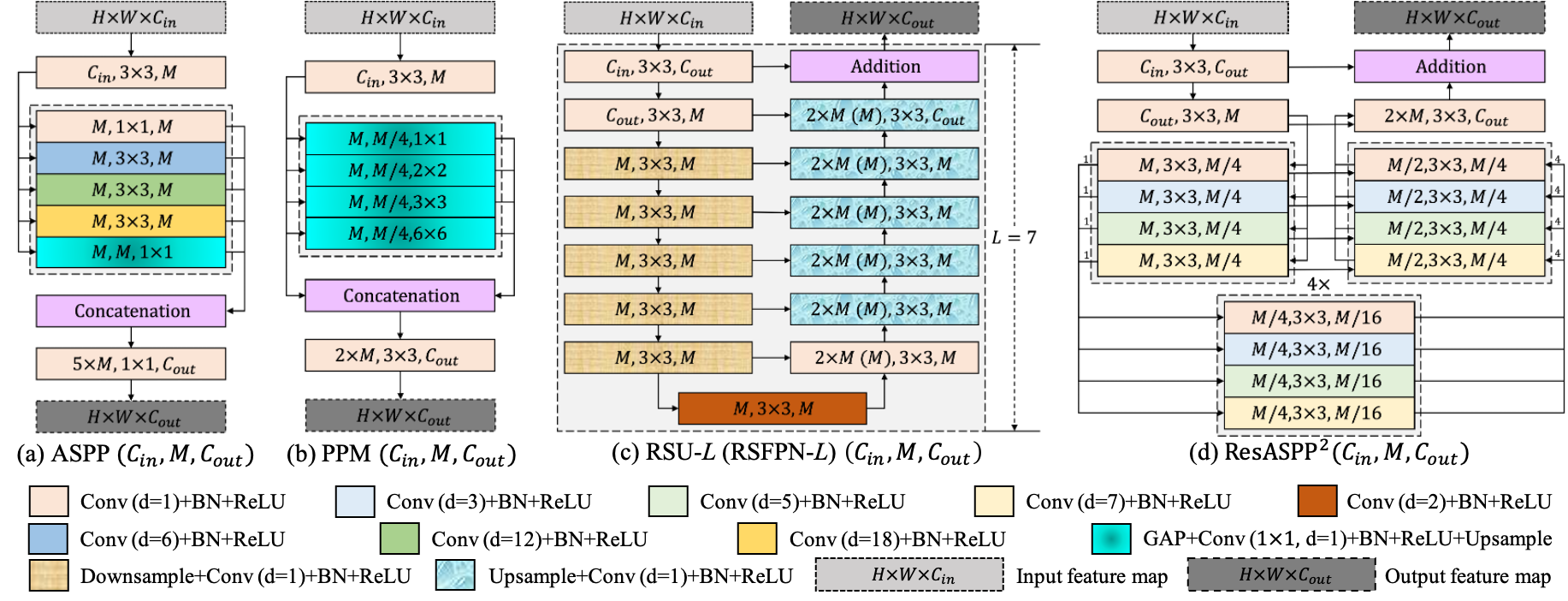}
\caption{Illustration of existing multi-scale feature fusion module and our proposed two-level Residual nested-ASPP module: (a) ASPP-like module, (b) PPM-like module, (c) RSU module and its extension RSFPN module, where $L$ is the number of layers in the encoder, (d) Our two-level Residual nested-ASPP module ResASPP$^{2}$.}
\label{fig:module comparison}
\end{figure*}

\section{Proposed Method}
First, we introduce our proposed Divide-and-Conquer Network and then describe the details of the two-level Residual nested-ASPP modules. 
Next we describe the Parallel Acceleration for DC-Net in detail. 
The training loss is described at the end of this section.
\subsection{Divide-and-Conquer Network}

The original use of the Divide-and-Conquer concept was to govern a nation, religion or country by first dividing it and then controlling and ruling it. Later, the same concept was applied to algorithms. The idea behind it is quite simple: divide a large or complex problem into smaller, simpler problems. Once the solutions to these smaller problems are obtained, they can be combined to solve the original problem. 

In this work, we propose a novel end-to-end network, named Divide-and-Conquer Network (DC-Net), by incorporating the concept of Divide-and-Conquer into the training process of salient object detection (SOD) networks. DC-Net divides the task of predicting saliency maps into $n$ subtasks, each responsible for predicting different semantic information of saliency maps. To achieve this, we supervise each stage of the encoder of every subtask with distinct auxiliary maps, while using the same encoder for each subtask. To reduce the GPU memory cost, we add an input convolutional layer with a kernel size of $3\times 3$ and a stride of 2 before the first stage of every subtask.

Here, we set $n$ to 2 to build our DC-Net as shown in Fig. \ref{fig:DC-Net}. DC-Net has 2 encoders \textbf{Encoder1} and \textbf{Encoder2}, each consisting of 4 stages (\textbf{En1\_1}, \textbf{En2\_1}, \textbf{En3\_1}, \textbf{En4\_1} and \textbf{En1\_2}, \textbf{En2\_2}, \textbf{En3\_2}, \textbf{En4\_2}), and a decoder consisting of 5 stages (\textbf{De1}, \textbf{De2}, \textbf{De3}, \textbf{De4}, \textbf{De5}). The input to each decoder stage (\textbf{De(N)}) is the concatenation of the output of \textbf{En(N)\_1}, \textbf{En(N)\_2}, and \textbf{De(N+1)}, where \textbf{N} is in $\{1, 2, 3, 4\}$, and the input to \textbf{De5} is the concatenation of the output of \textbf{En4\_1} and \textbf{En4\_2} after downsampling. Our method generates all side output predicted maps \textbf{Sup1\_1}, \textbf{Sup2\_1}, \textbf{Sup3\_1}, \textbf{Sup4\_1}, \textbf{Sup1\_2}, \textbf{Sup2\_2}, \textbf{Sup3\_2}, \textbf{Sup4\_2}, \textbf{Sup1}, \textbf{Sup2}, \textbf{Sup3}, \textbf{Sup4}, and \textbf{Sup5} from all encoder and decoder stages similar to HED \cite{xie2015holistically} by passing their outputs through a $3\times 3$ convolutional layer and a sigmoid function, and then upsampling the logits of these maps to the input image size. We choose edge maps with width 4 (only for the pixels salient in the saliency maps) and location maps, as shown in Fig. \ref{fig:auxiliary maps} (i) and (c), as target maps for two subtasks, which learn edge and location representations of salient objects respectively. The saliency map is used to supervise each decoder stage. We choose the output predicted map \textbf{Sup1} as our final saliency map.

\subsection{Two-Level Residual Nested-ASPP Modules}

For tasks such as salient object detection or other pixel-level tasks, both local and global semantic information are crucial. Local semantic information can be learned by shallow layers of the network, while global information depends on the size of the receptive field of the network. The most typical methods of enlarging the receptive field are as follows. The first one is to use the atrous convolution proposed by Deeplab \cite{chen2017deeplab}. The atrous convolution can obtain a larger receptive field than ordinary convolution without sacrificing image resolution. The atrous spatial pyramid pooling (ASPP) (as shown in Fig. \ref{fig:module comparison} (a)) consisting of atrous convolutions with different dilation rates obtains output feature maps with rich semantic information by fusing multi-scale features. The second is to use global average pooling (GAP) of different sizes similar to the pyramid pooling modules (PPM) (as shown in Fig. \ref{fig:module comparison} (b)) proposed by PSPNet \cite{zhao2017pyramid} to obtain prior information of different scales and different sub-regions, and then concatenate them with the original feature map, and after another convolutional layer, the output feature map with global semantic information is obtained. RSU\cite{qin2020u2} and RSFPN (which we modify based on RSU) is to continuously obtain feature maps of different scales through downsampling, then upsample and aggregate low-level and high-level with different scales step by step like U-Net \cite{ronneberger2015u} and FPN \cite{lin2017feature} (as shown in Fig. \ref{fig:module comparison} (c)). Their shortcomings are also obvious. ASPP has the disadvantage of sparse pixel sampling. PPM requires the original feature map to have a good feature representation. U-Net and FPN sacrifice the high resolution of the feature map in the process of downsampling and require more convolutional layers to obtain a larger receptive field, which leads to a large model size.

Inspired by the methods mentioned above, we propose a novel two-level Residual nested-ASPP module, \textbf{ResASPP$^{2}$}, to capture compact multi-scale features. In theory, ResASPP$^{2}$ can be extended to ResASPP$^{n}$, where the exponent $n$ can be set as an arbitrary positive integer. We set $n$ to 2 as it balances performance and efficiency mostly. The structure of ResASPP$^{2}(C_{in},M,C_{out})$ is shown in Fig. \ref{fig:module comparison} (d), where $C_{in}$, $C_{out}$ denote input and output channels and $M$ denotes the number of channels in the internal layers of ResASPP$^{2}$. Our ResASPP$^{2}$ mainly consists of three components:

\begin{enumerate}[1)]
    \item an input convolution layer, which transforms the input feature map $\mathrm{x}$ $(H\times W\times C_{in})$ to an intermediate map $\mathcal{F}(\mathrm{x})$ with channel of $C_{out}$ which contains local feature.
    
    \item Different from the dilation rate setting \{1, 6, 12, 18\} in ASPP, we set the dilation rate of each layer of ResASPP$^{2}$ to \{1, 3, 5, 7\} to obtain more compact pixel sampling. After two-level nested-ASPP, a feature map $\mathcal{ASPP}^{2}(\mathcal{F}(\mathrm{x}))$ with channel of $C_{out}$ is obtained, which has a larger receptive field and more multi-scale contextual information than ASPP, under a smaller dilation rate. $\mathcal{ASPP}^{2}$ represents the part of Fig. \ref{fig:module comparison} (d) other than the input convolutional layer.

    \begin{figure}[t]
    \centering
    \includegraphics[scale=1.4]{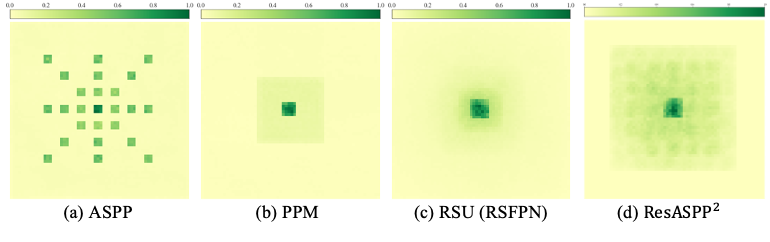}
    \caption{Comparison of the effective receptive field (ERF) of ASPP-like module, PPM-like module, RSU (RSFPN) module and our ResASPP$^{2}$ module.}
    \label{fig:erf}
    \vspace{-0.5em}
    \end{figure}

    Fig. \ref{fig:erf} presents a comparison of the effective receptive field (ERF) \cite{luo2016understanding} of various modules, including a single ASPP-like module, PPM-like module, RSU (RSFPN) module, and our proposed ResASPP$^{2}$ module. ResASPP$^{2}$ module outperforms other modules with the largest ERF and more compact ERF than ASPP. While RSU (RSFPN) module achieves its largest receptive field on the feature map with the lowest resolution after continuous downsampling, the decay of the gradient signal is exponential, resulting in a smaller ERF of its feature map obtained from the last layer after continuous upsampling and convolution. According to \cite{ding2022scaling}, the ERF is proportional to $\mathcal{O}(K\sqrt{L})$, where $K$ is the kernel size and $L$ is the depth (i.e., number of layers). Due to the fewer layers of ResASPP$^{2}$, the decay of the receptive field is negligible. Although RSU (RSFPN) has a larger largest receptive field than ResASPP$^{2}$, the ERF of its feature map obtained from the last layer is smaller than that of ResASPP$^{2}$. Furthermore, ResASPP$^{2}$ maintains high resolution of the feature maps all the time, while RSU (RSFPN) loses detail information in the process of continuous downsampling.

    \item a residual connection is used to fuse local features with multi-scale features through addition: $\mathcal{F}(\mathrm{x})+\mathcal{ASPP}^{2}(\mathcal{F}(\mathrm{x}))$.
    
\end{enumerate}


\subsection{Parallel Acceleration}
\begin{figure*}[t]
    \centering
    \includegraphics[width=\textwidth]{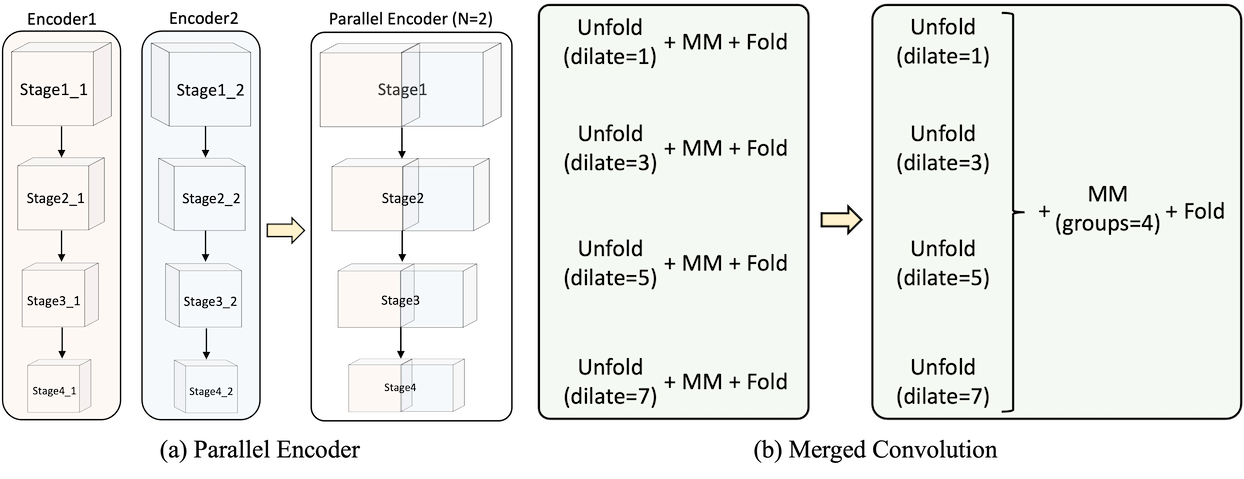}
    \caption{Illustration of the parallel encoder and merged convolution. 'MM' means Matrix Multiplication. A convolution operation can be separated as three parts: an unfold operation, a matrix multiplication, and a fold operation.}
    \label{fig:PA}
\end{figure*}

One advantage of the Divide-and-Conquer approach is its potential for parallel computing, which can improve the efficiency of the network. As shown in Fig. \ref{fig:PA}, the two identical encoders responsible for different subtasks can perform forward propagation simultaneously. To fully exploit this potential, we merge these two encoders into a single encoder with the same structure (\textbf{Parallel Encoder}) by reparameterizing operations such as convolutional layers, linear layers, matrix dot products, and layer normalization. Additionally, our ResASPP$^{2}$ module is accelerated by a proposed operation called \textbf{Merged Convolution}, which merges parallel convolutions with the same kernel size and output size. This allows for the computation of multiple parallel convolutions in a single step, reducing the total number of operations and accelerating the processing speed.

\subsection{Loss Function}
Our training loss function is defined as follows:
\begin{equation}
\mathnormal{L}=\sum_{e=1}^{E}(\mathnormal{w}_{1}^{(e)}\mathnormal{l}_{1}^{(e)}+\mathnormal{w}_{2}^{(e)}\mathnormal{l}_{2}^{(e)})+\sum_{d=1}^{D}\mathnormal{w}^{(d)}\mathnormal{l}^{(d)}
\end{equation}
In this equation, $\mathnormal{l}_{1}^{(e)}$ and $\mathnormal{l}_{2}^{(e)}$ are the losses of the side output auxiliary maps of \textbf{En($e$)\_1} and \textbf{En($e$)\_2} (referred to as \textbf{Sup($e$)\_1} and \textbf{Sup($e$)\_2} in Fig. \ref{fig:DC-Net}), where $e$ denotes the $e_{th}$ encoder out of a total of $E$ stages. $\mathnormal{l}^{(d)}$ is the loss of the side output saliency maps of \textbf{De($d$)}, where $d$ denotes the $d_{th}$ decoder out of a total of $D$ stages. The weights of each loss term are denoted by $\mathnormal{w}{1}^{(e)}$, $\mathnormal{w}{2}^{(e)}$, and $\mathnormal{w}^{(d)}$, respectively.

For each term $\mathnormal{l}_{1}$ and $\mathnormal{l}_{2}$, we use the standard binary cross entropy to calculate the loss:
\begin{equation}
\begin{aligned}
\mathnormal{l}_{bce}=-\sum_{(x,y)}^{(H,W)}&[g(x,y)log(p(x,y)) \\
&+(1-g(x,y))log(1-p(x,y))]
\end{aligned}
\label{eq:bce}
\end{equation}
where $(x,y)$ is the pixel coordinates and $(H,W)$ is the height and width of the image. $g(x,y)$ and $p(x,y)$ denote the pixel values of the ground truth and predicted probability map respectively. For each term $\mathnormal{l}$, to take the global structure of the image into account,
in addition to using the standard binary cross entropy, we also use IoU to calculate the loss:
\begin{equation}
\begin{aligned}
\mathnormal{l}_{iou}=1-\frac{\sum_{(x,y)}^{(H,W)}[g(x,y)p(x,y)]}{\sum_{(x,y)}^{(H,W)}[g(x,y)+p(x,y)-g(x,y)p(x,y)]}
\end{aligned}
\end{equation}
where the notations are the same as Eq. \ref{eq:bce}. The goal of our training process is to minimize the overall loss $\mathnormal{L}$. 

\section{Experiments}
\subsection{Implementation Details}
In the training process, we use data augmentation including horizontal flip, random crop, and multi-scale input images.
Two pretrained ResNet-34 \cite{he2016deep} and Swin-B \cite{liu2021swin} are used as the encoders of our DC-Net-R and DC-Net-S respectively, and other parameters are randomly initialized. The loss weights $\mathnormal{w}_{1}^{e}$, $\mathnormal{w}_{2}^{e}$ and $\mathnormal{w}^{d}$ are all set to 1. Stochastic gradient descent (SGD) optimizer with momentum \cite{rumelhart1986learning} is used to train our network and its learning rate is set to 0.01 for LR datasets (ResNet-34), 0.001 for HR datasets (ResNet-34), and 0.001 for LR datasets (Swin-B), other hyperparameters including momentum and weight decay are set to 0.9 and 0.0001. We set the batch size to 32 for LR datasets (ResNet-34), 4 for HR datasets (ResNet-34), and 8 for LR datasets (Swin-B) and train the network for around 60k iterations until the loss converges. In addition, we use apex\footnote{https://github.com/NVIDIA/apex} and fp16 to accelerate the training process. During inference, each image is first resized to $352\times 352$ for LR datasets (ResNet-34), $1024\times 1024$ for HR datasets (ResNet-34), and $384\times 384$ for LR datasets (Swin-B). Our network is implemented based on PyTorch \cite{paszke2019pytorch}. Both training and testing and other experiments are conducted on a single RTX 6000 GPU (24GB memory).

\subsection{Parallel Acceleration Details}
We directly implement the Merged Convolution with PyTorch without modifying the underlying code written in C language. Using the Merged Convolution and Parallel Encoder for training can result in large memory costs and low efficiency, therefore, we use Training-DC-Net and Inference-DC-Net in training and inference phase respectively. The Training-DC-Net does not merge any operation, and the Inference-DC-Net uses Merged Convolution and Parallel Encoder. The parameters are copied from Training-DC-Net to Inference-DC-Net based on specific rules before inference.

\subsection{Evaluation Metrics}
To provide relatively comprehensive and unbiased evaluation of the quality of those output probability maps against the ground truth, nine different metrics including (1) Precision-Recall (PR) curves, (2) F-measure curves, (3) maximal F-measure ($maxF_{\beta} \uparrow$) \cite{achanta2009frequency}, (4) Mean Absolute Error ($MAE \downarrow$), 
(5) weighted F-measure ($F_{\beta}^{w} \uparrow$) \cite{margolin2014evaluate}, 
(6) structural measure ($S_{\alpha} \uparrow$) \cite{fan2017structure},
(7) mean enhanced alignment measure ($E_{\phi}^{m} \uparrow$) \cite{fan2018enhanced}, (8) relax human correction efforts ($HCE_{\gamma}$) \cite{qin2022highly}, (9) mean boundary accuracy ($mBA$) \cite{cheng2020cascadepsp} are used:
\begin{enumerate}[(1)]
    \item \textbf{PR Curve} is generated using a collection of precision-recall pairs. When given a saliency probability map, its precision and recall scores are evaluated by comparing its thresholded binary mask with the actual ground truth mask. The precision and recall scores for the entire dataset are obtained by averaging the scores of individual saliency maps. By varying the thresholds between 0 and 255, a group of average precision-recall pairs for the dataset can be obtained.

    \item \textbf{F-measure Curve} draws the change of F-measure under different thresholds. For different thresholds between 0 and 255, the F-measure value of each dataset is obtained by averaging the F-measure value computed by comparing thresholded binary mask of each saliency probability map and its corresponding ground truth mask.

    \item \textbf{F-measure} ($F_{\beta}$) is a weighted harmonic mean of precision and recall: \begin{equation}
    F_{\beta}=\frac{(1+\beta^{2})\times Precision\times Recall}
    {\beta^{2}\times Precision+Recall}
    \end{equation}
    We set the $\beta^{2}$ to 0.3 similar to previous works \cite{achanta2009frequency,qin2020u2}. $F_{\beta}$ has different values for different thresholds between 0 and 255, and we report the maximum $F_{\beta}$ ($maxF_{\beta}$) for each dataset.

    \item \textbf{MAE} is the Mean Absolute Error which is calculated by averaging pixel-wise difference between the predicted saliency map (P) and the ground truth mask (G):
    \begin{equation}
    MAE=\frac{1}{H\times W}\sum^{H}_{x=1}\sum^{W}_{y=1}|P(x,y)-G(x,y)|
    \end{equation}

    \item \textbf{weighted F-measure} ($F_{\beta}^{w}$) is proposed to overcome the possible unfair comparison caused by interpolation flaw, dependency flaw and equal-imporance flaw \cite{liu2020picanet}:
    \begin{equation}
    F_{\beta}^{w}=\frac{(1+\beta^{2})\times Precision^{w}\times Recall^{w}}
    {\beta^{2}\times Precision^{w}+Recall^{w}}
    \end{equation}
    We set $\beta^{2}$ to 1.0 as suggested in \cite{borji2015salient}, and the weights ($w$) is different to each pixel according to its specific location and neighborhood information.

    \item \textbf{S-measure} ($S_{\alpha}$) is used to evaluate the object-aware ($S_{o}$) and region-aware ($S_{r}$) structural similarity, which is computed as:
    \begin{equation}
    S_{\alpha}=(1-m)S_{r}+m S_{o}
    \end{equation}
    We set $\alpha$ to 0.5 as suggested in \cite{fan2017structure}.

    \item \textbf{E-measure} ($E_{\phi}^{m}$) considers the local pixel values with the image-level mean value in one term, which can be defined as:
    \begin{equation}
    E_{\phi}=\frac{1}{H\times W}\sum^{H}_{x=1}\sum^{W}_{y=1}\phi(x,y)
    \end{equation}
    , where $\phi=f(\xi)$ is defined as the enhanced alignment matrix, $\xi$ is defined as an alignment matrix, and $f(x)=\frac{1}{4}(1+x)^{2}$ is a simple and effective function.
    We report mean E-measure ($E_{\phi}^{m}$) for each dataset.

    \item \textbf{relax HCE} ($HCE_{\gamma}$) aims to estimate the amount of human efforts needed to correct erroneous predictions and meet specific accuracy standards in practical scenarios, which can be defined as:
    \begin{equation}
    HCE_{\gamma}=compute\_HCE(FN',FP',TP,epsilon)
    \end{equation}
    We set $\gamma$ to 5 and $epsilon$ to 2.0 as suggested in \cite{qin2022highly}.

    \item \textbf{mBA} is used to evaluate the boundary quality, and \cite{kim2022revisiting} shows that mBA itself cannot measure the performance of saliency detection, rather it only measures the quality of boundary itself.
    
\end{enumerate}

\subsection{Ablation Study}
\textbf{Ablation on Auxiliary Maps:}
In the auxiliary maps ablation, the goal is to find the most effective auxiliary map combination of subtasks. As shown in Table \ref{tab:aux}, we take the case of no subtask as the baseline, and we find that the performance of DC-Net-R is worse when the auxiliary maps all are saliency maps. We believe that predicting the saliency map is a difficult task, and the subtask should be simple, which may be the reason for the poor performance. Using the body and detail maps proposed in \cite{wei2020label} as auxiliary maps yields a performance comparable to the baseline.
Multi-value maps are more challenging than binary maps, making them unsuitable as subtasks. If we assume that predicting the saliency map involves a two-step process, where the first step is predicting the background pixel value as 0 and the second step is predicting the foreground pixel value as 1, then predicting the location map containing the location information of salient objects completes the first step, which is a simple binary prediction subtask. The edge map is a commonly used auxiliary map, and we have observe that the width of the edge pixel can impact the performance of the network. Our hypothesis is that a moderate width of the edge pixel can help the network focus more on the edges and avoid introducing excessive non-edge information.

\begin{table}[h]
    \centering
    \scriptsize
    \caption{Results of ablation study on auxiliary maps. The table compares the results when encoder1 and encoder2 are supervised by different auxiliary maps including saliency, body, detail, edge1, edge2, edge3, edge4, edge5 and location maps as shown in Fig. \ref{fig:auxiliary maps}. \textcolor{cyan}{Cyan} means the auxiliary maps that our DC-Net adopts.
    }
    \setlength{\tabcolsep}{.2mm}{
    \begin{tabular}{cc|ccccc|ccccc}
    \hline
    \multicolumn{2}{c}{\textbf{Auxiliary Map}}\vline&\multicolumn{5}{c}{\textbf{DUTS-TE}}\vline&\multicolumn{5}{c}{\textbf{HKU-IS}}\\
    Encoder1&Encoder2&$F_{\beta}^{w}$&$MAE$&$maxF_{\beta}$&$S_{\alpha}$&$E_{\phi}^{m}$&$F_{\beta}^{w}$&$MAE$&$maxF_{\beta}$&$S_{\alpha}$&$E_{\phi}^{m}$\\
    \hline
    -&-&.838&.040&.891&.888&.917&.902&.028&.940&.921&.950\\
    Saliency&Saliency&.829&.040&.887&.885&.915&.901&.029&.939&.921&.950\\
    Body&Detail&.837&.038&.891&.888&.917&.902&.029&.940&.920&.949\\
    Edge1&Location&.845&.037&.894&.891&.921&.905&.028&.941&.921&.951\\
    Edge2&Location&.845&.036&.896&.893&.923&.905&.028&.941&.922&.952\\
    Edge3&Location&.847&.036&.897&.893&.923&.905&.028&\textbf{.942}&.921&.951\\
    \cellcolor{cyan}Edge4&\cellcolor{cyan}Location&\textbf{.852}&\textbf{.035}&\textbf{.899}&\textbf{.896}&\textbf{.927}&\textbf{.909}&\textbf{.027}&\textbf{.942}&\textbf{.924}&\textbf{.954}\\
    Edge5&Location&.845&.036&.895&.892&.922&.905&.028&.941&.922&.951\\
    \hline
    
    \end{tabular}
    \label{tab:aux}}
    \vspace{-1.0em}
\end{table}

\textbf{Ablation on Modules:}
In the module ablation, the goal is to validate the effectiveness of our newly designed two-level Residual nested-ASPP module (ResASPP$^{2}$). 
Specifically, we fix the encoder part and the combination of subtasks (Edge4+Location) and replace each stage of the decoder with other modules in Fig. \ref{fig:module comparison}, including ASPP-like modules, PPM-like modules, RSU modules, and RSFPN modules. The module parameters $C_{in}$, $M$, and $C_{out}$ of each stage of different modules are the same.

Table \ref{tab:modules} shows the model size, FPS, and performance on DUTS-TE, HKU-IS datasets of DC-Net using different modules. Compared with RSU and RSFPN, our ResASPP$^{2}$ has a smaller model size when the FPS is competitive with them, and achieves better results on the datasets. Compared with the traditional two multi-scale contextual modules ASPP-like module and PPM-like module, ResASPP$^{2}$ greatly improves the performance on the datasets. Therefore, we believe that our newly designed ResASPP$^{2}$ can achieve better results than other modules in this salient object detection task.

\begin{table}[h]
    \centering
    \scriptsize
    \caption{Results of ablation study on modules. The structure of ASPP-like module, PPM-like module, RSU module, RSFPN module and ResASPP$^{2}$ module are shown in Fig. \ref{fig:module comparison}. \textcolor{cyan}{Cyan} means the module that our DC-Net adopts to the decoder.}
    \setlength{\tabcolsep}{.2mm}{
    \begin{tabular}{c|c|c|ccccc|ccccc}
    \hline
    \multirow{2}*{\textbf{Module}}&\multirow{2}*{\tabincell{c}{\textbf{Size}\\(MB)}}&\multirow{2}*{\textbf{FPS}}&\multicolumn{5}{c}{\textbf{DUTS-TE}}\vline&\multicolumn{5}{c}{\textbf{HKU-IS}}\\
    &&&$F_{\beta}^{w}$&$MAE$&$maxF_{\beta}$&$S_{\alpha}$&$E_{\phi}^{m}$&$F_{\beta}^{w}$&$MAE$&$maxF_{\beta}$&$S_{\alpha}$&$E_{\phi}^{m}$\\
    \hline
    ASPP \cite{chen2017deeplab}&269.3&66&.826&.040&.892&.878&.905&.893&.032&.938&.913&.939\\
    PPM \cite{zhao2017pyramid}&266.5&77&.830&.039&.885&.886&.915&.905&.028&.939&.922&.952\\
    RSU \cite{qin2020u2}&425.3&61&.842&.038&.894&.892&.920&.906&.028&.941&.923&.952\\
    RSFPN&374.1&63&.844&.038&.895&.894&.923&.906&\textbf{.027}&\textbf{.942}&\textbf{.924}&.952\\
    \hline
    \cellcolor{cyan}ResASPP$^{2}$&356.3&60&\textbf{.852}&\textbf{.035}&\textbf{.899}&\textbf{.896}&\textbf{.927}&\textbf{.909}&\textbf{.027}&\textbf{.942}&\textbf{.924}&\textbf{.954}\\
    \hline
    \end{tabular}
    \label{tab:modules}}
    \vspace{-1.0em}
\end{table}


\textbf{Ablation on Parallel Acceleration:}
As shown in Table \ref{tab:pa}, the ablation study on parallel acceleration compares the time costs of DC-Net-R with and without acceleration of encoder or ResASPP$^{2}$. Training-DC-Net and Inference-DC-Net have the lowest time costs in the training phase and inference phase, respectively. As we can see, the accelerated encoder and ResASPP$^{2}$ are 5 (21) ms and 9 (13) ms faster for DC-Net-R (DC-Net-S), respectively, for a total of 14 (34) ms faster.

\begin{table}[h]
    \centering
    \scriptsize
    \caption{Results of ablation study on parallel acceleration. 
    \CheckmarkBold and \XSolidBrush denote with and without acceleration respectively. \textcolor{cyan}{Cyan} and \textcolor{magenta}{Magenta} denote Training-DC-Net and Inference-DC-Net respectively. The batch sizes of training phase here are 12 (DC-Net-R) and 4 (DC-Net-S).}
    \setlength{\tabcolsep}{.22mm}{
    \begin{tabular}{cc|ccc|ccc}
    \hline
    \multicolumn{2}{c}{\textbf{Parallel Acceleration}}\vline&\multicolumn{3}{c}{\textbf{DC-Net-R}}\vline&\multicolumn{3}{c}{\textbf{DC-Net-S}}\\
    Encoder&ResASPP$^{2}$&\tabincell{c}{Training\\time (ms)}&\tabincell{c}{Inference\\time (ms)}&\tabincell{c}{Max\\batch size}&\tabincell{c}{Training\\time (ms)}&\tabincell{c}{Inference\\time (ms)}&\tabincell{c}{Max\\batch size}\\
    \hline
    \cellcolor{cyan}\XSolidBrush&\cellcolor{cyan}\XSolidBrush&\textbf{430}&31&\textbf{47}&\textbf{553}&69&\textbf{8}\\
    \CheckmarkBold&\XSolidBrush&2055&26&46&682&48&7\\
    \XSolidBrush&\CheckmarkBold&665&22&24&619&56&6\\
    \cellcolor{magenta}\CheckmarkBold&\cellcolor{magenta}\CheckmarkBold&2314&\textbf{17}&24&756&\textbf{35}&5\\
    \hline
    \end{tabular}
    \label{tab:pa}}
    \vspace{-1.0em}
\end{table}

\textbf{Ablation on Fusion Ways and Number of Encoders:}
\begin{table*}[t]
    \centering
    \scriptsize
    \caption{Results of ablation study on fusion ways and number of encoders. \CheckmarkBold and \XSolidBrush denote with and without auxiliary map combination Edge4$+$Location. 'A' denotes addition and 'C' denotes concatenation.}
    \resizebox{\textwidth}{!}{
    \begin{tabular}{c|c|c|c|c|ccccc|ccccc}
    \hline
    \multirow{2}*{\tabincell{c}{\textbf{Fusion}\\\textbf{Way}}}&\multirow{2}*{\tabincell{c}{\textbf{Number of}\\\textbf{Encoders}}}&\multirow{2}*{\tabincell{c}{\textbf{Auxiliary}\\\textbf{Map}}}&\multirow{2}*{\tabincell{c}{\textbf{Size}\\(MB)}}&\multirow{2}*{\textbf{FPS}}&\multicolumn{5}{c}{\textbf{DUTS-TE}}\vline&\multicolumn{5}{c}{\textbf{HKU-IS}}\\
    &&&&&$F_{\beta}^{w}$&$MAE$&$maxF_{\beta}$&$S_{\alpha}$&$E_{\phi}^{m}$&$F_{\beta}^{w}$&$MAE$&$maxF_{\beta}$&$S_{\alpha}$&$E_{\phi}^{m}$\\
    \hline
    A&1&\XSolidBrush&132.3&62&.836&.040&.890&.888&.916&.902&.029&.939&.920&.950\\
    A&2&\XSolidBrush&211.4&61&.838&.039&.892&.887&.917&.903&.029&.940&.921&.951\\
    A&3&\XSolidBrush&296.5&59&.842&.038&.894&.891&.919&.906&.028&.941&.923&.952\\
    A&4&\XSolidBrush&391.8&57&\textbf{.846}&\textbf{.036}&\textbf{.896}&\textbf{.892}&\textbf{.923}&\textbf{.909}&\textbf{.027}&\textbf{.943}&\textbf{.924}&\textbf{.954}\\
    \hline
    A&2&\CheckmarkBold&211.4&61&.846&.036&.897&.893&.923&.905&.028&.940&.922&.952\\
    C&2&\CheckmarkBold&356.3&60&\textbf{.852}&\textbf{.035}&\textbf{.899}&\textbf{.896}&\textbf{.927}&\textbf{.909}&\textbf{.027}&\textbf{.942}&\textbf{.924}&\textbf{.954}\\
    \hline
    \end{tabular}
    \label{tab:noe}}
    \vspace{-1.0em}
\end{table*}
In the ablation study on auxiliary maps, we demonstrate that supervising the model with an inappropriate combination of auxiliary maps can lead to a decrease in model performance, while a reasonable combination can increase model performance. In this ablation study, our goal is to prove the following idea: more encoders more effective. Therefore, finding a reasonable combination of auxiliary maps for more encoders is necessary. Additionally, we compare the effects of two different feature fusion ways. Considering that models using concatenation for feature fusion will have more parameters and computational complexity compared to the addition-based fusion way, we conduct an ablation study on the number of encoders using the addition-based fusion way. As shown in Table \ref{tab:noe}, with an increase in the number of encoders, the model performance improves and the number of parameters also increases. Model efficiency remains relatively stable due to the utilization of our \textbf{Parallel Encoder}.

By comparing the results in the $2_{nd}$ row and the $5_{th}$ row, we once again demonstrate that supervising the model with a reasonable combination of auxiliary maps can enhance model performance. We find that using concatenation for feature fusion performs better than addition, when not considering model size, we choose concatenation as the feature fusion way for DC-Net.

\subsection{Experiments on Low-Resolution Saliency Detection Datasets}
\subsubsection{Datasets}
\textbf{Training dataset:} DUTS dataset \cite{wang2017learning} is the largest and most frequently used training dataset for salient object detection currently. DUTS can be separated as a training dataset \textbf{DUTS-TR} and \textbf{DUTS-TE}, and we train our network on \textbf{DUTS-TR}, which contains 10553 images in total.

\textbf{Evaluation datasets:} We evaluate our network on five frequently used benchmark datasets including: \textbf{DUTS-TE} \cite{wang2017learning} with 5019 images, \textbf{DUT-OMRON} \cite{yang2013saliency} with 5168 images, \textbf{HKU-IS} \cite{li2016visual} with 4447 images, \textbf{ECSSD} \cite{yan2013hierarchical} with 1000 images, \textbf{PASCAL-S} \cite{li2014secrets} with 850 images.
In addition, we also measure the model performance on the challenging \textbf{SOC} (Salient Object in Clutter) test dataset \cite{fan2022salient} to show the generalization performance of our network in different scenarios.

\begin{table*}[!t]
    \centering
    \scriptsize
    \caption{Comparison of our method and 21 SOTA methods on DUTS-TE, DUT-OMRON, HKU-IS, ECSSD, and PASCAL-S in terms of $F_{\beta}^{w}$ ($\uparrow$), $MAE$ ($\downarrow$), $maxF_{\beta}$ ($\uparrow$), $S_{\alpha}$ ($\uparrow$) and $E_{\phi}^{m}$ ($\uparrow$). \textcolor{red}{Red}, \textcolor{green}{Green} and \textcolor{blue}{Blue} indicate the best, second best, and third best performance. The superscript of each score is the corresponding ranking. $'-'$ means missing data.}\label{tab:quantitative}
    \renewcommand{\arraystretch}{1.0}
    \setlength\tabcolsep{1.0pt}
    \resizebox{\textwidth}{!}{
    \begin{tabular}{r|cccc|lllll|lllll|lllll|lllll|lllll}
        \hline
        \multirow{2}*{\textbf{Method}}&\multirow{2}*{\textbf{Backbone}}&\multirow{2}*{\tabincell{c}{\textbf{Size}\\(MB)}}&\multirow{2}*{\tabincell{c}{\textbf{Input}\\\textbf{Size}}}&\multirow{2}*{\textbf{FPS}}&\multicolumn{5}{c}{\textbf{DUTS-TE}(5019)}\vline&\multicolumn{5}{c}{\textbf{DUT-OMRON}(5168)}\vline&\multicolumn{5}{c}{\textbf{HKU-IS}(4447)}\vline&\multicolumn{5}{c}{\textbf{ECSSD}(1000)}\vline&\multicolumn{5}{c}{\textbf{PASCAL-S}(850)}\\
        &&&&&$F_{\beta}^{w}$&$MAE$&$maxF_{\beta}$&$S_{\alpha}$&$E_{\phi}^{m}$&$F_{\beta}^{w}$&$MAE$&$maxF_{\beta}$&$S_{\alpha}$&$E_{\phi}^{m}$&$F_{\beta}^{w}$&$MAE$&$maxF_{\beta}$&$S_{\alpha}$&$E_{\phi}^{m}$&$F_{\beta}^{w}$&$MAE$&$maxF_{\beta}$&$S_{\alpha}$&$E_{\phi}^{m}$&$F_{\beta}^{w}$&$MAE$&$maxF_{\beta}$&$S_{\alpha}$&$E_{\phi}^{m}$\\
        \hline
        \multicolumn{30}{c}{\textbf{Convolution-Based Methods}}\\
        \hline
        \textbf{PoolNet-R}$_{19}$ \cite{liu2019simple}&ResNet-50&278.5&300$\times$400&54&.817$^{10}$&\textcolor{blue}{.037}$^{3}$&.889$^{5}$&.887$^{6}$&.910$^{9}$&.725$^{13}$&.054$^{4}$&.805$^{13}$&.831$^{13}$&.848$^{12}$&.888$^{9}$&.030$^{4}$&.936$^{5}$&\textcolor{blue}{.919}$^{3}$&.945$^{7}$&.904$^{9}$&.035$^{4}$&\textcolor{blue}{.949}$^{3}$&.926$^{4}$&.945$^{5}$&.809$^{7}$&.065$^{7}$&\textcolor{red}{.879}$^{1}$&\textcolor{green}{.865}$^{2}$&.896$^{5}$\\
        \textbf{SCRN}$_{19}$ \cite{wu2019stacked}&ResNet-50&101.4&352$\times$352&38&.803$^{15}$&.040$^{6}$&.888$^{6}$&.885$^{7}$&.900$^{13}$&.720$^{14}$&.056$^{6}$&.811$^{10}$&.837$^{8}$&.848$^{12}$&.876$^{13}$&.034$^{8}$&.934$^{7}$&.916$^{6}$&.935$^{12}$&.900$^{11}$&.037$^{5}$&\textcolor{green}{.950}$^{2}$&\textcolor{blue}{.927}$^{3}$&.939$^{10}$&.807$^{9}$&.063$^{5}$&\textcolor{green}{.877}$^{2}$&\textcolor{red}{.869}$^{1}$&.892$^{9}$\\
        \textbf{AFNet}$_{19}$ \cite{feng2019attentive}&VGG-16&133.6&224$\times$224&-&.785$^{17}$&.046$^{10}$&.863$^{14}$&.867$^{13}$&.893$^{17}$&.717$^{16}$&.057$^{7}$&.797$^{15}$&.826$^{14}$&.846$^{14}$&.869$^{15}$&.036$^{9}$&.922$^{13}$&.905$^{10}$&.934$^{13}$&.886$^{14}$&.042$^{7}$&.935$^{11}$&.913$^{11}$&.935$^{12}$&.797$^{11}$&.070$^{10}$&.863$^{12}$&.849$^{12}$&.883$^{12}$\\
        \textbf{BASNet}$_{19}$ \cite{qin2019basnet}&ResNet-34&348.5&256$\times$256&88&.803$^{15}$&.048$^{11}$&.859$^{15}$&.866$^{14}$&.896$^{16}$&.751$^{5}$&.056$^{6}$&.805$^{13}$&.836$^{9}$&.865$^{5}$&.889$^{8}$&.032$^{6}$&.928$^{10}$&.909$^{9}$&.943$^{9}$&.904$^{9}$&.037$^{5}$&.942$^{8}$&.916$^{10}$&.943$^{7}$&.793$^{14}$&.076$^{13}$&.854$^{15}$&.838$^{16}$&.879$^{15}$\\
        \textbf{BANet}$_{19}$ \cite{su2019selectivity}&ResNet-50&203.2&400$\times$300&-&.811$^{12}$&.040$^{6}$&.872$^{11}$&.879$^{10}$&.913$^{7}$&.736$^{10}$&.059$^{8}$&.803$^{14}$&.832$^{12}$&.870$^{2}$&.886$^{11}$&.032$^{6}$&.931$^{9}$&.913$^{8}$&.946$^{6}$&.908$^{8}$&.035$^{4}$&.945$^{6}$&.924$^{6}$&\textcolor{blue}{.948}$^{3}$&.802$^{10}$&.070$^{10}$&.864$^{11}$&.852$^{11}$&.891$^{10}$\\
        \textbf{EGNet-R}$_{19}$ \cite{zhao2019egnet}&ResNet-50&447.1&352$\times$352&53&.816$^{11}$&.039$^{5}$&.889$^{5}$&.887$^{6}$&.907$^{11}$&.738$^{9}$&\textcolor{blue}{.053}$^{3}$&.815$^{7}$&.841$^{4}$&.857$^{9}$&.887$^{10}$&.031$^{5}$&.935$^{6}$&.918$^{4}$&.944$^{8}$&.903$^{10}$&.037$^{5}$&.947$^{5}$&.925$^{5}$&.943$^{7}$&.795$^{12}$&.074$^{12}$&.865$^{10}$&.852$^{11}$&.881$^{14}$\\
        \textbf{CPD-R}$_{19}$ \cite{wu2019cascaded}&ResNet-50&192.0&352$\times$352&37&.795$^{16}$&.043$^{8}$&.865$^{13}$&.869$^{12}$&.898$^{14}$&.719$^{15}$&.056$^{6}$&.797$^{15}$&.825$^{15}$&.847$^{13}$&.875$^{14}$&.034$^{8}$&.925$^{12}$&.905$^{10}$&.938$^{10}$&.898$^{12}$&.037$^{5}$&.939$^{9}$&.918$^{9}$&.942$^{8}$&.794$^{13}$&.071$^{11}$&.859$^{14}$&.848$^{13}$&.882$^{13}$\\
        \textbf{U$^{2}$-Net}$_{20}$ \cite{qin2020u2}&RSU&176.3&320$\times$320&41&.804$^{14}$&.045$^{9}$&.873$^{10}$&.874$^{11}$&.897$^{15}$&\textcolor{blue}{.757}$^{3}$&.054$^{4}$&\textcolor{blue}{.823}$^{3}$&\textcolor{green}{.847}$^{2}$&\textcolor{blue}{.867}$^{3}$&.889$^{8}$&.031$^{5}$&.935$^{6}$&.916$^{6}$&.943$^{9}$&.910$^{7}$&\textcolor{green}{.033}$^{2}$&\textcolor{red}{.951}$^{1}$&\textcolor{green}{.928}$^{2}$&.947$^{4}$&.792$^{15}$&.074$^{12}$&.859$^{14}$&.844$^{14}$&.873$^{16}$\\
        \textbf{RASNet}$_{20}$ \cite{chen2020reverse}&VGG-16&98.6&352$\times$352&83&.827$^{6}$&\textcolor{blue}{.037}$^{3}$&.886$^{7}$&.884$^{8}$&.920$^{4}$&.743$^{8}$&.055$^{5}$&.815$^{7}$&.836$^{9}$&.866$^{4}$&.894$^{6}$&.030$^{4}$&.933$^{8}$&.915$^{7}$&.950$^{4}$&.913$^{4}$&\textcolor{blue}{.034}$^{3}$&.948$^{4}$&.925$^{5}$&\textcolor{green}{.950}$^{2}$&-&-&-&-&-\\
        \textbf{F$^{3}$Net}$_{20}$ \cite{wei2020f3net}&ResNet-50&102.5&352$\times$352&63&.835$^{5}$&\textcolor{green}{.035}$^{2}$&.891$^{4}$&.888$^{5}$&.920$^{4}$&.747$^{7}$&\textcolor{blue}{.053}$^{3}$&.813$^{8}$&.838$^{7}$&.864$^{6}$&.900$^{4}$&\textcolor{green}{.028}$^{2}$&.937$^{4}$&.917$^{5}$&\textcolor{blue}{.952}$^{3}$&.912$^{5}$&\textcolor{green}{.033}$^{2}$&.945$^{6}$&.924$^{6}$&\textcolor{blue}{.948}$^{3}$&.816$^{4}$&\textcolor{blue}{.061}$^{3}$&.871$^{6}$&.861$^{5}$&.898$^{4}$\\
        \textbf{MINet-R}$_{20}$ \cite{pang2020multi}&ResNet-50&650.0&320$\times$320&42&.825$^{7}$&\textcolor{blue}{.037}$^{3}$&.884$^{8}$&.884$^{8}$&.917$^{6}$&.738$^{9}$&.056$^{6}$&.810$^{11}$&.833$^{11}$&.860$^{7}$&.897$^{5}$&\textcolor{blue}{.029}$^{3}$&.935$^{6}$&\textcolor{blue}{.919}$^{3}$&\textcolor{blue}{.952}$^{3}$&.911$^{6}$&\textcolor{green}{.033}$^{2}$&.947$^{5}$&.925$^{5}$&\textcolor{green}{.950}$^{2}$&.809$^{7}$&.064$^{6}$&.866$^{9}$&.856$^{10}$&.896$^{5}$\\
        \textbf{CAGNet-R}$_{20}$ \cite{mohammadi2020cagnet}&ResNet-50&199.8&480$\times$480&-&.817$^{10}$&.040$^{6}$&.867$^{12}$&.864$^{15}$&.909$^{10}$&.729$^{12}$&.054$^{4}$&.791$^{16}$&.814$^{16}$&.855$^{10}$&.893$^{7}$&.030$^{4}$&.926$^{11}$&.904$^{11}$&.946$^{6}$&.903$^{10}$&.037$^{5}$&.937$^{10}$&.907$^{12}$&.941$^{9}$&.808$^{8}$&.066$^{8}$&.860$^{13}$&.842$^{15}$&.893$^{8}$\\
        \textbf{GateNet-R}$_{20}$ \cite{zhao2020suppress}&ResNet-50&514.9&384$\times$384&65&.809$^{13}$&.040$^{6}$&.888$^{6}$&.885$^{7}$&.906$^{12}$&.729$^{12}$&.055$^{5}$&.818$^{6}$&.838$^{7}$&.855$^{10}$&.880$^{12}$&.033$^{7}$&.933$^{8}$&.915$^{7}$&.937$^{11}$&.894$^{13}$&.040$^{6}$&.945$^{6}$&.920$^{8}$&.936$^{11}$&.797$^{11}$&.067$^{9}$&.869$^{8}$&.858$^{8}$&.886$^{11}$\\
        \textbf{ITSD-R}$_{20}$ \cite{zhou2020interactive}&ResNet-50&106.2&288$\times$288&52&.824$^{8}$&.041$^{7}$&.883$^{9}$&.885$^{7}$&.913$^{7}$&.750$^{6}$&.061$^{9}$&.821$^{4}$&.840$^{5}$&.865$^{5}$&.894$^{6}$&.031$^{5}$&.934$^{7}$&.917$^{5}$&.947$^{5}$&.910$^{7}$&\textcolor{blue}{.034}$^{3}$&.947$^{5}$&.925$^{5}$&.947$^{4}$&.812$^{6}$&.066$^{8}$&.870$^{7}$&.859$^{7}$&.894$^{7}$\\
        \textbf{GCPANet}$_{20}$ \cite{chen2020global}&ResNet-50&268.6&288$\times$288&61&.821$^{9}$&.038$^{4}$&.888$^{6}$&\textcolor{blue}{.891}$^{3}$&.911$^{8}$&.734$^{11}$&.056$^{6}$&.812$^{9}$&.839$^{6}$&.853$^{11}$&.889$^{8}$&.031$^{5}$&\textcolor{blue}{.938}$^{3}$&\textcolor{green}{.920}$^{2}$&.944$^{8}$&.903$^{10}$&.035$^{4}$&.948$^{4}$&\textcolor{blue}{.927}$^{3}$&.944$^{6}$&.808$^{8}$&.062$^{4}$&.869$^{8}$&\textcolor{blue}{.864}$^{3}$&.895$^{6}$\\
        \textbf{LDF}$_{20}$ \cite{wei2020label}&ResNet-50&100.9&352$\times$352&63&\textcolor{green}{.845}$^{2}$&\textcolor{red}{.034}$^{1}$&\textcolor{green}{.897}$^{2}$&\textcolor{green}{.892}$^{2}$&\textcolor{green}{.925}$^{2}$&.752$^{4}$&\textcolor{green}{.052}$^{2}$&.820$^{5}$&.839$^{6}$&.865$^{5}$&\textcolor{green}{.904}$^{2}$&\textcolor{green}{.028}$^{2}$&\textcolor{green}{.939}$^{2}$&\textcolor{blue}{.919}$^{3}$&\textcolor{green}{.953}$^{2}$&\textcolor{blue}{.915}$^{3}$&\textcolor{blue}{.034}$^{3}$&\textcolor{green}{.950}$^{2}$&.924$^{6}$&\textcolor{blue}{.948}$^{3}$&\textcolor{green}{.822}$^{2}$&\textcolor{green}{.060}$^{2}$&.874$^{5}$&.863$^{4}$&\textcolor{red}{.903}$^{1}$\\
        \textbf{ICON-R}$_{21}$ \cite{zhuge2022salient}&ResNet-50&132.8&352$\times$352&53&.837$^{4}$&\textcolor{blue}{.037}$^{3}$&\textcolor{blue}{.892}$^{3}$&.889$^{4}$&\textcolor{blue}{.924}$^{3}$&\textcolor{green}{.761}$^{2}$&.057$^{7}$&\textcolor{green}{.825}$^{2}$&\textcolor{blue}{.844}$^{3}$&\textcolor{red}{.876}$^{1}$&\textcolor{blue}{.902}$^{3}$&\textcolor{blue}{.029}$^{3}$&\textcolor{green}{.939}$^{2}$&\textcolor{green}{.920}$^{2}$&\textcolor{green}{.953}$^{2}$&\textcolor{red}{.918}$^{1}$&\textcolor{red}{.032}$^{1}$&\textcolor{green}{.950}$^{2}$&\textcolor{red}{.929}$^{1}$&\textcolor{red}{.954}$^{1}$&\textcolor{blue}{.818}$^{3}$&.064$^{6}$&\textcolor{blue}{.876}$^{3}$&.861$^{5}$&\textcolor{blue}{.899}$^{3}$\\
        \textbf{RCSB}$_{22}$ \cite{ke2022recursive}&ResNet-50&107.4&256$\times$256&21&\textcolor{blue}{.840}$^{3}$&\textcolor{green}{.035}$^{2}$&.889$^{5}$&.881$^{9}$&.919$^{5}$&.752$^{4}$&\textcolor{red}{.049}$^{1}$&.809$^{12}$&.835$^{10}$&.858$^{8}$&\textcolor{red}{.909}$^{1}$&\textcolor{red}{.027}$^{1}$&\textcolor{blue}{.938}$^{3}$&\textcolor{blue}{.919}$^{3}$&\textcolor{red}{.954}$^{1}$&\textcolor{green}{.916}$^{2}$&\textcolor{blue}{.034}$^{3}$&.944$^{7}$&.922$^{7}$&\textcolor{green}{.950}$^{2}$&\textcolor{red}{.826}$^{1}$&\textcolor{red}{.059}$^{1}$&.875$^{4}$&.860$^{6}$&\textcolor{green}{.902}$^{2}$\\
        \hline 
        \rowcolor{gray!20}\textbf{DC-Net-R (Ours-R)}&ResNet-34&356.3&352$\times$352&60&\textcolor{red}{.852}$^{1}$&\textcolor{green}{.035}$^{2}$&\textcolor{red}{.899}$^{1}$&\textcolor{red}{.896}$^{1}$&\textcolor{red}{.927}$^{1}$&\textcolor{red}{.772}$^{1}$&\textcolor{blue}{.053}$^{3}$&\textcolor{red}{.827}$^{1}$&\textcolor{red}{.849}$^{1}$&\textcolor{red}{.876}$^{1}$&\textcolor{red}{.909}$^{1}$&\textcolor{red}{.027}$^{1}$&\textcolor{red}{.942}$^{1}$&\textcolor{red}{.924}$^{1}$&\textcolor{red}{.954}$^{1}$&.913$^{4}$&\textcolor{blue}{.034}$^{3}$&\textcolor{blue}{.949}$^{3}$&.924$^{6}$&.945$^{5}$&.814$^{5}$&.066$^{8}$&.874$^{5}$&.857$^{9}$&.892$^{9}$\\
        \hline
        \multicolumn{30}{c}{\textbf{Self-Attention-Based Methods}}\\
        \hline
        \textbf{VST}$_{21}$ \cite{liu2021visual}&T2T-ViT$_{t}$-14&178.4&224$\times$224&35&.828$^{4}$&.037$^{4}$&.890$^{4}$&.896$^{4}$&.919$^{4}$&.755$^{4}$&\textcolor{blue}{.058}$^{3}$&.824$^{4}$&.850$^{4}$&.871$^{4}$&.897$^{4}$&.029$^{4}$&.942$^{4}$&.928$^{4}$&.952$^{4}$&.910$^{4}$&\textcolor{blue}{.033}$^{3}$&.951$^{4}$&.932$^{4}$&.951$^{4}$&\textcolor{blue}{.816}$^{3}$&.061$^{4}$&.875$^{4}$&.872$^{4}$&.902$^{4}$\\
        \textbf{ICON-S}$_{21}$ \cite{zhuge2022salient}&Swin-B&383.5&384$\times$384&29&\textcolor{green}{.886}$^{2}$&\textcolor{green}{.025}$^{2}$&\textcolor{green}{.920}$^{2}$&\textcolor{green}{.917}$^{2}$&\textcolor{red}{.954}$^{1}$&\textcolor{green}{.804}$^{2}$&\textcolor{green}{.043}$^{2}$&\textcolor{green}{.855}$^{2}$&\textcolor{green}{.869}$^{2}$&\textcolor{red}{.900}$^{1}$&\textcolor{green}{.925}$^{2}$&\textcolor{green}{.022}$^{2}$&\textcolor{green}{.951}$^{2}$&\textcolor{green}{.935}$^{2}$&\textcolor{red}{.968}$^{1}$&\textcolor{green}{.936}$^{2}$&\textcolor{red}{.023}$^{1}$&\textcolor{green}{.961}$^{2}$&\textcolor{green}{.941}$^{2}$&\textcolor{red}{.966}$^{1}$&\textcolor{red}{.854}$^{1}$&\textcolor{red}{.048}$^{1}$&\textcolor{green}{.896}$^{2}$&\textcolor{green}{.885}$^{2}$&\textcolor{red}{.924}$^{1}$\\
        \textbf{SelfReformer}$_{22}$ \cite{yun2022selfreformer}&PVT&366.7&224$\times$224&21&\textcolor{blue}{.872}$^{3}$&\textcolor{blue}{.027}$^{3}$&\textcolor{blue}{.916}$^{3}$&\textcolor{blue}{.911}$^{3}$&\textcolor{blue}{.943}$^{3}$&\textcolor{blue}{.784}$^{3}$&\textcolor{green}{.043}$^{2}$&\textcolor{blue}{.837}$^{3}$&\textcolor{blue}{.861}$^{3}$&\textcolor{blue}{.884}$^{3}$&\textcolor{blue}{.915}$^{3}$&\textcolor{blue}{.024}$^{3}$&\textcolor{blue}{.947}$^{3}$&\textcolor{blue}{.931}$^{3}$&\textcolor{blue}{.960}$^{3}$&\textcolor{blue}{.926}$^{3}$&\textcolor{green}{.027}$^{2}$&\textcolor{blue}{.958}$^{3}$&\textcolor{blue}{.936}$^{3}$&\textcolor{blue}{.957}$^{3}$&\textcolor{green}{.848}$^{2}$&\textcolor{blue}{.051}$^{3}$&\textcolor{blue}{.894}$^{3}$&\textcolor{blue}{.881}$^{3}$&\textcolor{green}{.919}$^{2}$\\
        \hline
        \rowcolor{gray!20}\textbf{DC-Net-S (Ours-S)}&Swin-B&1495.0&384$\times$384&29&\textcolor{red}{.895}$^{1}$&\textcolor{red}{.023}$^{1}$&\textcolor{red}{.930}$^{1}$&\textcolor{red}{.925}$^{1}$&\textcolor{green}{.952}$^{2}$&\textcolor{red}{.809}$^{1}$&\textcolor{red}{.039}$^{1}$&\textcolor{red}{.857}$^{1}$&\textcolor{red}{.875}$^{1}$&\textcolor{green}{.898}$^{2}$&\textcolor{red}{.929}$^{1}$&\textcolor{red}{.021}$^{1}$&\textcolor{red}{.956}$^{1}$&\textcolor{red}{.941}$^{1}$&\textcolor{green}{.966}$^{2}$&\textcolor{red}{.941}$^{1}$&\textcolor{red}{.023}$^{1}$&\textcolor{red}{.966}$^{1}$&\textcolor{red}{.947}$^{1}$&\textcolor{green}{.965}$^{2}$&\textcolor{red}{.854}$^{1}$&\textcolor{green}{.049}$^{2}$&\textcolor{red}{.899}$^{1}$&\textcolor{red}{.887}$^{1}$&\textcolor{blue}{.917}$^{3}$\\
        \hline
    \end{tabular}
    }
    \vspace{-1.0em}
\end{table*}

\begin{figure*}[h]
    \centering
    \includegraphics[width=\textwidth]{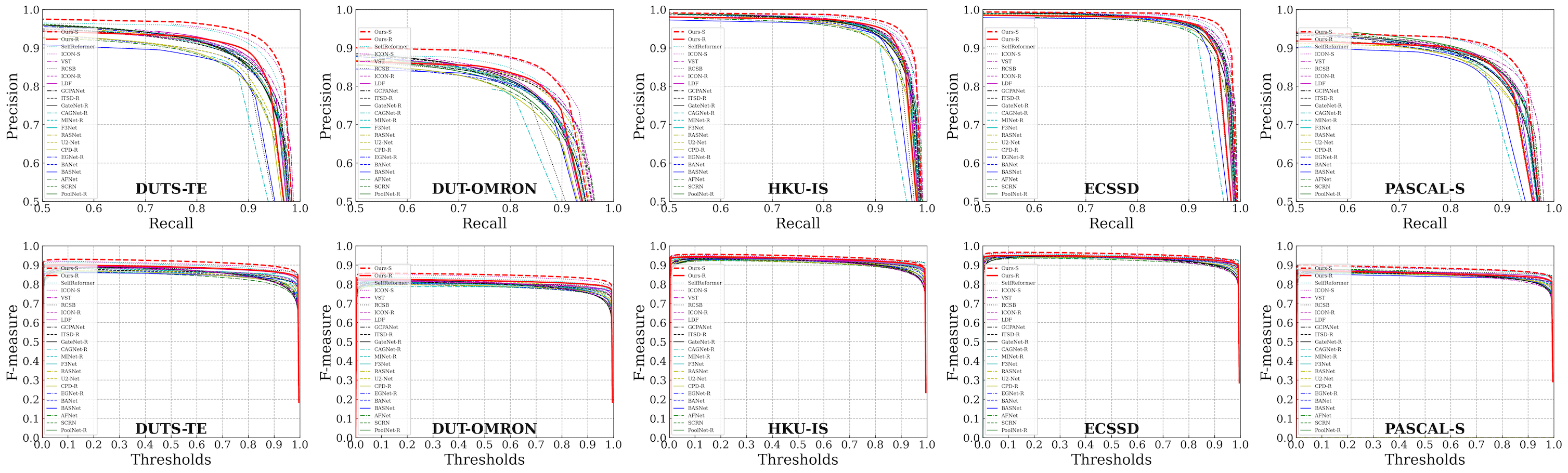}
    \caption{First row: Precision-Recall Curves comparison on five low-resolution saliency benchmark datasets. Second row: F-measure Curves comparison on five low-resolution saliency benchmark datasets.}
    \label{fig:pr_fm}
\end{figure*}

\subsubsection{Comparison with State-of-the-arts}
\subsubsubsection{Dataset-Based Analysis}
We compare our DC-Net-R with 18 recent four years state-of-the-art convolution-based methods including one \textbf{RSU} based model: \textbf{U$^{2}$-Net}; one \textbf{ResNet-34} based model: \textbf{BASNet}; two \textbf{VGG-16} \cite{simonyan2014very} based model: \textbf{AFNet}, \textbf{RASNet}; six \textbf{ResNet-50} based model: \textbf{SCRN}, \textbf{BANet}, \textbf{F$^{3}$Net}, \textbf{GCPANet}, \textbf{LDF}, \textbf{RCSB}; for the other eight methods \textbf{PoolNet}, \textbf{EGNet}, \textbf{CPD}, \textbf{MINet}, \textbf{CAGNet}, \textbf{GateNet}, \textbf{ITSD}, \textbf{ICON}, we selected their better models based on \textbf{ResNet} or \textbf{VGG} for comparison. We also compare our DC-Net-S with 3 state-of-the-art self-attention-based methods including one \textbf{T2T-ViT$_{t}$-14} based model: \textbf{VST}; one \textbf{Swin-B} based model: \textbf{ICON-S}; one \textbf{PVT} based model: \textbf{SelfReformer}. For a fair comparison, we use the salient object detection results provided by the authors, and the same inference code is used to test the FPS of methods.

\textbf{Quantitative Comparison:}
Table \ref{tab:quantitative} compares five evaluation metrics including $maxF_{\beta}$, $MAE$, $F_{\beta}^{w}$, $S_{\alpha}$ and $E_{\phi}^{m}$ of our proposed method with others. As we can see, our DC-Net performs against the existing methods across almost all five traditional benchmark datasets in terms of nearly all evaluation metrics.
Fig. \ref{fig:pr_fm} illustrates the precision-recall curves and F-measure curves which are consistent with Table \ref{tab:quantitative}. The two red lines belonging to the proposed method are higher than the other curves, which further shows the effectiveness of prior knowledge and large ERF.

\textbf{Qualitative Comparison:}
\begin{figure*}[!t]
    \centering
    \includegraphics[width=\textwidth]{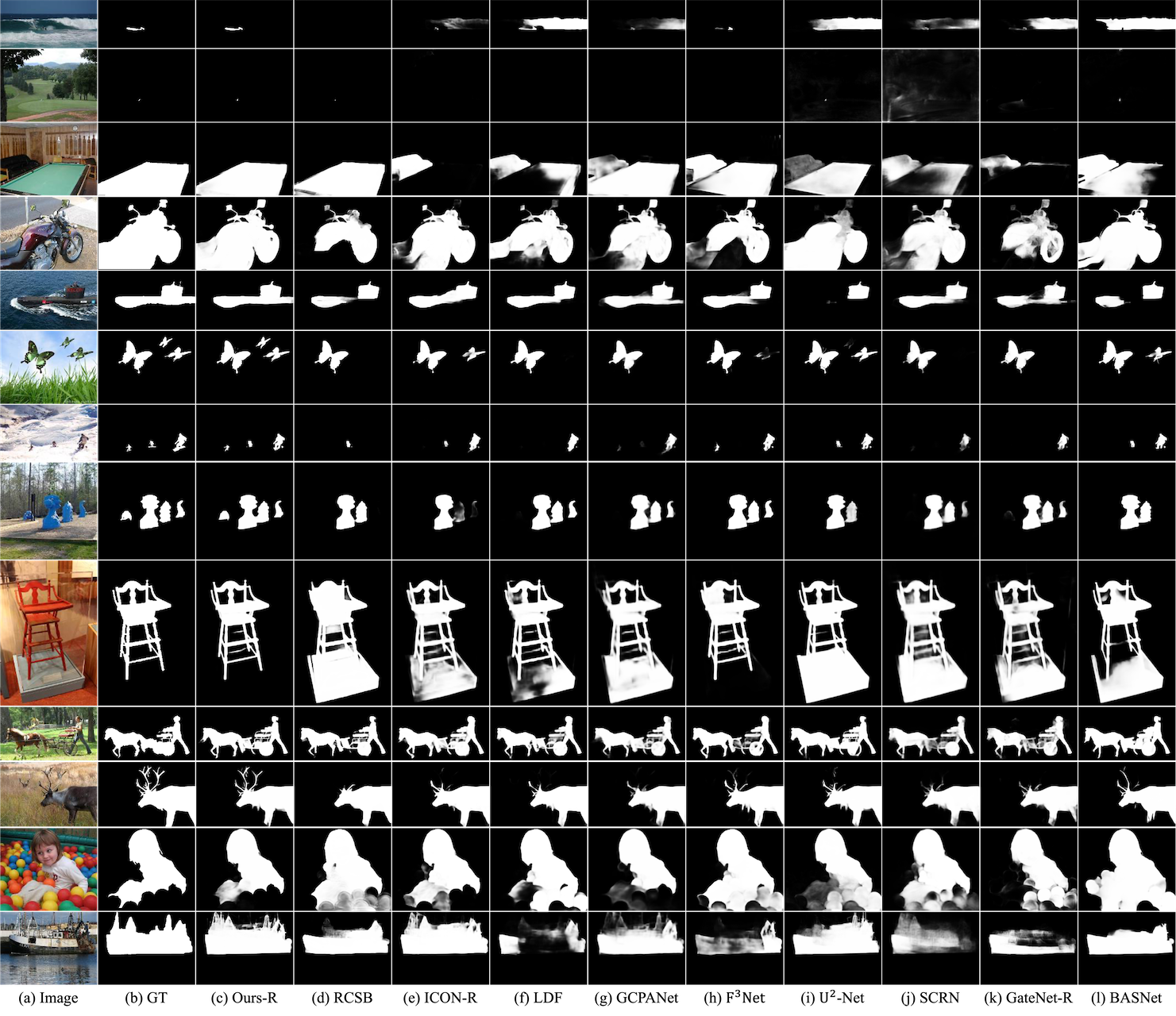}
    \caption{Low-resolution dataset-based qualitative comparison of the proposed method with nine other SOTA methods: (a) Image, (b) GT, (c) Ours-R, (d) RCSB, (e) ICON-R, (f) LDF, (g) GCPANet, (h) F$^{3}$Net, (i) U$^{2}$-Net, (j) SCRN, (k) GateNet-R, (l) BASNet.}
    \label{fig:qualitative}
\end{figure*}
Fig. \ref{fig:qualitative} shows the sample results of our method and other eight best-performing methods and the method with the first best FPS in Table \ref{tab:quantitative}, which intuitively demonstrates the promising performance of our method in different scenarios.

The $1_{st}$ and $2_{nd}$ rows of Fig. \ref{fig:qualitative} show the results for small and hidden objects. Among all methods, only our DC-Net can accurately find the location of the object in the $1_{st}$ row image and segment it. The $3_{rd}$, $4_{th}$ and $5_{th}$ rows show the results for large objects that extend to the edges of the image and our method can accurately segment the salient objects with high confidence. The $6_{th}$, $7_{th}$ and $8_{th}$ rows show the scenario where there are multiple objects of the same categories that are near or far. We can find that our DC-Net is able to segment all objects accurately, while other methods miss one or more objects. The $9_{th}$, $10_{th}$ and $11_{th}$ rows represent the scenario of objects with thin structures. As we can observe, our DC-Net can accurately segment even better than the chair part of the ground truth of the $10_{th}$ row. The $11_{th}$ and $12_{th}$ rows show the scenario where the image has a complex background. In this case, most of the time it is difficult for humans to distinguish the foreground from the background accurately. Compared with other methods, our method shows a better performance.

\textbf{Failure Cases:}
\begin{figure*}[!t]
    \centering
    \includegraphics[width=\textwidth]{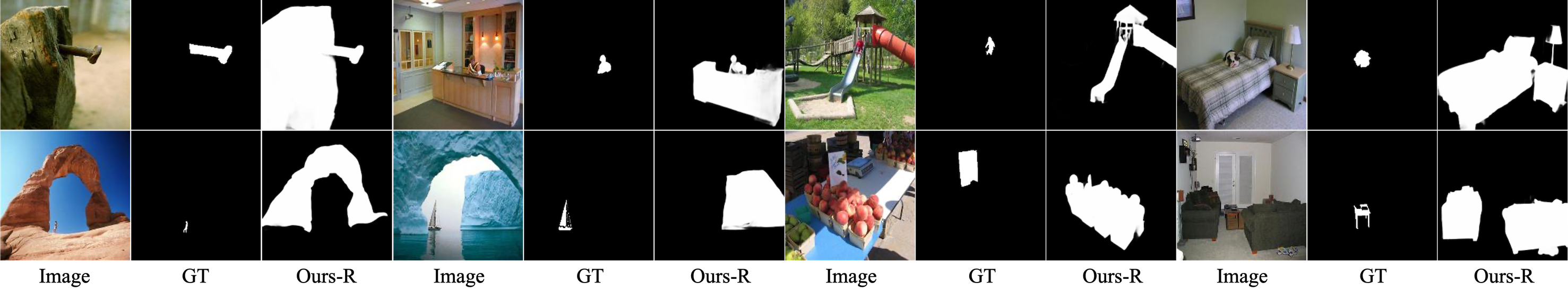}
    \caption{Failure cases of dataset-based analysis.}
    \label{fig:failurecase}
\end{figure*}
\begin{table*}[!t]
    \centering
    \scriptsize
    \caption{Comparison of our method and 18 SOTA methods on SOC test dataset in terms of $maxF_{\beta}$ ($\uparrow$), $MAE$ ($\downarrow$), $F_{\beta}^{w}$ ($\uparrow$), $S_{\alpha}$ ($\uparrow$) and $E_{\phi}^{m}$ ($\uparrow$). \textcolor{red}{Red}, \textcolor{green}{Green} and \textcolor{blue}{Blue} indicate the best, second best and third best performance. The superscript of each score is the corresponding ranking.}\label{tab:soc_quantitative}
    \renewcommand{\arraystretch}{1.0}
    \setlength\tabcolsep{1.0pt}
    \resizebox{\textwidth}{!}{
    \begin{tabular}{c|c|llllllllllllllllll|l}
        \hline
        \multirow{2}*{\textbf{Attr}}&\multirow{2}*{\textbf{Metrics}}&Amulet&DSS&NLDF&SRM&BMPM&C2SNet&DGRL&R$^{3}$Net&RANet&AFNet&BASNet&CPD&EGNet&PoolNet&SCRN&BANet&MINet&PiCANet&DC-Net-R\\
        &&\cite{zhang2017amulet}&\cite{hou2017deeply}&\cite{luo2017non}&\cite{wang2017stagewise}&\cite{zhang2018bi}&\cite{li2018contour}&\cite{wang2018detect}&\cite{deng2018r3net}&\cite{chen2018reverse}&\cite{feng2019attentive}&\cite{qin2019basnet}&\cite{wu2019cascaded}&\cite{zhao2019egnet}&\cite{liu2019simple}&\cite{wu2019stacked}&\cite{su2019selectivity}&\cite{pang2020multi}&\cite{liu2020picanet}&(Ours-R)\\
        \hline
        \multirow{5}*{\textbf{AC}}
        &$maxF_{\beta} \uparrow$&.752$^{14}$&.755$^{13}$&.751$^{15}$&.804$^{7}$&.791$^{10}$&.752$^{14}$&.785$^{11}$&.756$^{12}$&.745$^{16}$&.801$^{8}$&.804$^{7}$&.811$^{5}$&\textcolor{green}{.822}$^{2}$&.801$^{8}$&.817$^{4}$&\textcolor{blue}{.819}$^{3}$&.808$^{6}$&.795$^{9}$&\textcolor{red}{.834}$^{1}$\\
        &$MAE \downarrow$&.120$^{16}$&.113$^{14}$&.119$^{15}$&.096$^{11}$&.098$^{12}$&.109$^{13}$&.081$^{4}$&.135$^{18}$&.132$^{17}$&.084$^{6}$&.083$^{5}$&.089$^{9}$&.085$^{7}$&.093$^{10}$&\textcolor{green}{.078}$^{2}$&.086$^{8}$&\textcolor{blue}{.079}$^{3}$&.093$^{10}$&\textcolor{red}{.076}$^{1}$\\
        &$F_{\beta}^{w} \uparrow$&.620$^{16}$&.629$^{15}$&.620$^{16}$&.690$^{11}$&.680$^{13}$&.647$^{14}$&.718$^{8}$&.593$^{18}$&.603$^{17}$&.712$^{10}$&.727$^{5}$&.721$^{7}$&\textcolor{blue}{.731}$^{3}$&.713$^{9}$&.723$^{6}$&\textcolor{green}{.739}$^{2}$&.730$^{4}$&.681$^{12}$&\textcolor{red}{.768}$^{1}$\\
        &$S_{\alpha} \uparrow$&.752$^{13}$&.753$^{12}$&.737$^{14}$&.791$^{9}$&.780$^{10}$&.755$^{11}$&.791$^{9}$&.713$^{15}$&.709$^{16}$&.796$^{6}$&.799$^{5}$&.799$^{5}$&\textcolor{blue}{.806}$^{3}$&.795$^{7}$&\textcolor{green}{.809}$^{2}$&\textcolor{blue}{.806}$^{3}$&.802$^{4}$&.793$^{8}$&\textcolor{red}{.824}$^{1}$\\
        &$E_{\phi}^{m} \uparrow$&.790$^{14}$&.787$^{15}$&.783$^{16}$&.824$^{10}$&.815$^{11}$&.806$^{13}$&.853$^{4}$&.752$^{18}$&.765$^{17}$&.852$^{5}$&.842$^{9}$&.852$^{5}$&\textcolor{blue}{.854}$^{3}$&.846$^{7}$&.848$^{6}$&\textcolor{green}{.858}$^{2}$&.843$^{8}$&.814$^{12}$&\textcolor{red}{.867}$^{1}$\\
        \hline
        \multirow{5}*{\textbf{BO}}
        &$maxF_{\beta} \uparrow$&.814$^{16}$&.813$^{17}$&.835$^{12}$&.853$^{11}$&.826$^{15}$&.863$^{10}$&.888$^{5}$&.782$^{18}$&.725$^{19}$&.874$^{7}$&.868$^{9}$&.895$^{4}$&.829$^{14}$&.831$^{13}$&\textcolor{red}{.921}$^{1}$&.879$^{6}$&\textcolor{blue}{.917}$^{3}$&\textcolor{green}{.919}$^{2}$&.872$^{8}$\\
        &$MAE \downarrow$&.334$^{12}$&.343$^{15}$&.341$^{14}$&.294$^{11}$&.292$^{10}$&.257$^{7}$&\textcolor{blue}{.207}$^{3}$&.432$^{17}$&.440$^{18}$&.236$^{5}$&.247$^{6}$&.236$^{5}$&.358$^{16}$&.339$^{13}$&.217$^{4}$&.261$^{8}$&\textcolor{red}{.175}$^{1}$&\textcolor{green}{.192}$^{2}$&.278$^{9}$\\
        &$F_{\beta}^{w} \uparrow$&.625$^{15}$&.628$^{14}$&.635$^{13}$&.679$^{12}$&.683$^{11}$&.739$^{8}$&\textcolor{blue}{.794}$^{3}$&.471$^{17}$&.469$^{18}$&.750$^{6}$&.749$^{7}$&.755$^{5}$&.602$^{16}$&.625$^{15}$&.784$^{4}$&.729$^{9}$&\textcolor{red}{.828}$^{1}$&\textcolor{green}{.805}$^{2}$&.699$^{10}$\\
        &$S_{\alpha} \uparrow$&.589$^{13}$&.577$^{16}$&.583$^{14}$&.628$^{11}$&.619$^{12}$&.667$^{7}$&.696$^{4}$&.455$^{18}$&.437$^{19}$&.671$^{6}$&.660$^{8}$&.679$^{5}$&.546$^{17}$&.578$^{15}$&\textcolor{blue}{.707}$^{3}$&.657$^{9}$&\textcolor{red}{.743}$^{1}$&\textcolor{green}{.739}$^{2}$&.637$^{10}$\\
        &$E_{\phi}^{m} \uparrow$&.566$^{14}$&.554$^{16}$&.556$^{15}$&.630$^{12}$&.635$^{11}$&.674$^{8}$&\textcolor{blue}{.736}$^{3}$&.435$^{18}$&.423$^{19}$&.710$^{5}$&.678$^{7}$&.699$^{6}$&.547$^{17}$&.572$^{13}$&.716$^{4}$&.663$^{9}$&\textcolor{red}{.769}$^{1}$&\textcolor{green}{.750}$^{2}$&.641$^{10}$\\
        \hline
        \multirow{5}*{\textbf{CL}}
        &$maxF_{\beta} \uparrow$&.781$^{11}$&.731$^{16}$&.727$^{17}$&.770$^{14}$&.771$^{13}$&.748$^{15}$&.785$^{9}$&.687$^{18}$&.681$^{19}$&.803$^{7}$&.792$^{8}$&.806$^{5}$&.782$^{10}$&.779$^{12}$&\textcolor{blue}{.811}$^{3}$&.808$^{4}$&\textcolor{green}{.822}$^{2}$&.805$^{6}$&\textcolor{red}{.830}$^{1}$\\
        &$MAE \downarrow$&.141$^{10}$&.153$^{12}$&.159$^{13}$&.134$^{8}$&.123$^{7}$&.144$^{11}$&.119$^{6}$&.182$^{14}$&.188$^{15}$&.119$^{6}$&.114$^{4}$&\textcolor{green}{.112}$^{2}$&.139$^{9}$&.134$^{8}$&\textcolor{blue}{.113}$^{3}$&.117$^{5}$&\textcolor{red}{.108}$^{1}$&.123$^{7}$&\textcolor{green}{.112}$^{2}$\\
        &$F_{\beta}^{w} \uparrow$&.663$^{13}$&.617$^{15}$&.614$^{16}$&.665$^{12}$&.678$^{10}$&.655$^{14}$&.714$^{6}$&.546$^{17}$&.542$^{18}$&.696$^{7}$&\textcolor{blue}{.724}$^{3}$&.719$^{4}$&.677$^{11}$&.681$^{9}$&.717$^{5}$&\textcolor{green}{.725}$^{2}$&.719$^{4}$&.691$^{8}$&\textcolor{red}{.746}$^{1}$\\
        &$S_{\alpha} \uparrow$&.763$^{10}$&.721$^{15}$&.713$^{16}$&.758$^{12}$&.76$^{11}$&.742$^{14}$&.769$^{8}$&.659$^{17}$&.633$^{18}$&.767$^{9}$&.773$^{7}$&.786$^{4}$&.757$^{13}$&.760$^{11}$&\textcolor{green}{.795}$^{2}$&.784$^{5}$&.783$^{6}$&\textcolor{blue}{.787}$^{3}$&\textcolor{red}{.798}$^{1}$\\
        &$E_{\phi}^{m} \uparrow$&.788$^{12}$&.763$^{14}$&.764$^{13}$&.792$^{10}$&.801$^{7}$&.789$^{11}$&\textcolor{green}{.824}$^{2}$&.709$^{16}$&.715$^{15}$&.802$^{6}$&.821$^{4}$&\textcolor{blue}{.823}$^{3}$&.789$^{11}$&.800$^{8}$&.819$^{5}$&\textcolor{green}{.824}$^{2}$&.819$^{5}$&.793$^{9}$&\textcolor{red}{.834}$^{1}$\\
        \hline
        \multirow{5}*{\textbf{HO}}
        &$maxF_{\beta} \uparrow$&.804$^{10}$&.789$^{14}$&.778$^{15}$&.800$^{11}$&.791$^{13}$&.771$^{16}$&.792$^{12}$&.766$^{17}$&.757$^{18}$&.814$^{9}$&.833$^{4}$&.826$^{7}$&.828$^{6}$&\textcolor{blue}{.836}$^{3}$&\textcolor{blue}{.836}$^{3}$&.831$^{5}$&\textcolor{red}{.840}$^{1}$&.819$^{8}$&\textcolor{green}{.838}$^{2}$\\
        &$MAE \downarrow$&.119$^{14}$&.124$^{16}$&.126$^{17}$&.115$^{12}$&.116$^{13}$&.123$^{15}$&.104$^{9}$&.135$^{18}$&.143$^{19}$&.102$^{8}$&.097$^{5}$&.098$^{6}$&.106$^{10}$&.100$^{7}$&.096$^{4}$&\textcolor{blue}{.094}$^{3}$&\textcolor{red}{.089}$^{1}$&.109$^{11}$&\textcolor{green}{.092}$^{2}$\\
        &$F_{\beta}^{w} \uparrow$&.688$^{12}$&.660$^{16}$&.661$^{15}$&.696$^{11}$&.684$^{13}$&.668$^{14}$&.722$^{8}$&.633$^{17}$&.626$^{18}$&.722$^{8}$&.751$^{4}$&.736$^{7}$&.720$^{9}$&.739$^{6}$&.743$^{5}$&\textcolor{blue}{.753}$^{3}$&\textcolor{green}{.759}$^{2}$&.703$^{10}$&\textcolor{red}{.761}$^{1}$\\
        &$S_{\alpha} \uparrow$&.790$^{13}$&.767$^{16}$&.755$^{17}$&.794$^{11}$&.781$^{14}$&.768$^{15}$&.791$^{12}$&.740$^{18}$&.713$^{19}$&.798$^{10}$&.803$^{8}$&.807$^{7}$&.802$^{9}$&.815$^{5}$&\textcolor{red}{.823}$^{1}$&\textcolor{blue}{.819}$^{3}$&\textcolor{green}{.821}$^{2}$&.809$^{6}$&.818$^{4}$\\
        &$E_{\phi}^{m} \uparrow$&.809$^{13}$&.796$^{16}$&.798$^{15}$&.819$^{10}$&.813$^{12}$&.805$^{14}$&.833$^{8}$&.781$^{17}$&.777$^{18}$&.833$^{8}$&.844$^{5}$&.838$^{7}$&.829$^{9}$&.845$^{4}$&.842$^{6}$&\textcolor{green}{.850}$^{2}$&\textcolor{red}{.858}$^{1}$&.817$^{11}$&\textcolor{blue}{.848}$^{3}$\\
        \hline
        \multirow{5}*{\textbf{MB}}
        &$maxF_{\beta} \uparrow$&.680$^{17}$&.717$^{14}$&.698$^{15}$&.746$^{9}$&.741$^{10}$&.692$^{16}$&.739$^{11}$&.674$^{18}$&.733$^{13}$&.763$^{7}$&\textcolor{green}{.792}$^{2}$&.734$^{12}$&.779$^{5}$&.765$^{6}$&\textcolor{red}{.801}$^{1}$&.783$^{4}$&.783$^{4}$&.750$^{8}$&\textcolor{blue}{.790}$^{3}$\\
        &$MAE \downarrow$&.142$^{14}$&.132$^{11}$&.138$^{12}$&.115$^{8}$&\textcolor{blue}{.105}$^{3}$&.128$^{10}$&.113$^{7}$&.160$^{15}$&.139$^{13}$&.111$^{6}$&.106$^{4}$&\textcolor{green}{.104}$^{2}$&.109$^{5}$&.121$^{9}$&\textcolor{red}{.100}$^{1}$&\textcolor{green}{.104}$^{2}$&\textcolor{blue}{.105}$^{3}$&\textcolor{red}{.100}$^{1}$&.109$^{5}$\\
        &$F_{\beta}^{w} \uparrow$&.561$^{15}$&.577$^{13}$&.551$^{16}$&.619$^{11}$&.651$^{6}$&.593$^{12}$&.655$^{5}$&.489$^{17}$&.576$^{14}$&.626$^{10}$&\textcolor{green}{.679}$^{2}$&.655$^{5}$&.649$^{7}$&.642$^{8}$&\textcolor{red}{.690}$^{1}$&.670$^{4}$&\textcolor{blue}{.676}$^{3}$&.636$^{9}$&\textcolor{blue}{.676}$^{3}$\\
        &$S_{\alpha} \uparrow$&.712$^{14}$&.719$^{13}$&.685$^{16}$&.742$^{11}$&.762$^{4}$&.719$^{13}$&.744$^{10}$&.657$^{17}$&.696$^{15}$&.734$^{12}$&.754$^{7}$&.753$^{8}$&.762$^{4}$&.751$^{9}$&\textcolor{red}{.792}$^{1}$&\textcolor{blue}{.764}$^{3}$&.761$^{5}$&\textcolor{green}{.775}$^{2}$&.757$^{6}$\\
        &$E_{\phi}^{m} \uparrow$&.738$^{15}$&.753$^{13}$&.739$^{14}$&.777$^{10}$&\textcolor{blue}{.812}$^{3}$&.777$^{10}$&\textcolor{red}{.823}$^{1}$&.697$^{16}$&.761$^{12}$&.762$^{11}$&.803$^{5}$&.809$^{4}$&.789$^{7}$&.779$^{9}$&\textcolor{green}{.816}$^{2}$&.803$^{5}$&.793$^{6}$&\textcolor{blue}{.812}$^{3}$&.787$^{8}$\\
        \hline
        \multirow{5}*{\textbf{OC}}
        &$maxF_{\beta} \uparrow$&.731$^{13}$&.722$^{15}$&.713$^{16}$&.747$^{11}$&.747$^{11}$&.728$^{14}$&.732$^{12}$&.674$^{18}$&.677$^{17}$&.775$^{5}$&.763$^{9}$&\textcolor{green}{.780}$^{2}$&.768$^{7}$&.771$^{6}$&\textcolor{blue}{.778}$^{3}$&.766$^{8}$&.776$^{4}$&.762$^{10}$&\textcolor{red}{.790}$^{1}$\\
        &$MAE \downarrow$&.143$^{13}$&.144$^{14}$&.149$^{15}$&.129$^{11}$&.119$^{9}$&.130$^{12}$&.116$^{7}$&.168$^{16}$&.169$^{17}$&\textcolor{blue}{.109}$^{3}$&.115$^{6}$&\textcolor{green}{.106}$^{2}$&.121$^{10}$&.118$^{8}$&.111$^{4}$&.112$^{5}$&\textcolor{red}{.102}$^{1}$&.119$^{9}$&\textcolor{red}{.102}$^{1}$\\
        &$F_{\beta}^{w} \uparrow$&.607$^{14}$&.595$^{15}$&.593$^{16}$&.63$^{12}$&.644$^{10}$&.622$^{13}$&.659$^{8}$&.520$^{18}$&.527$^{17}$&\textcolor{blue}{.680}$^{3}$&.672$^{7}$&.679$^{4}$&.658$^{9}$&.659$^{8}$&.673$^{6}$&.677$^{5}$&\textcolor{green}{.686}$^{2}$&.637$^{11}$&\textcolor{red}{.708}$^{1}$\\
        &$S_{\alpha} \uparrow$&.735$^{14}$&.719$^{15}$&.709$^{16}$&.749$^{11}$&.752$^{9}$&.738$^{13}$&.747$^{12}$&.653$^{17}$&.641$^{18}$&.771$^{4}$&.750$^{10}$&\textcolor{blue}{.773}$^{3}$&.754$^{8}$&.756$^{7}$&\textcolor{green}{.775}$^{2}$&.766$^{5}$&.771$^{4}$&.765$^{6}$&\textcolor{red}{.787}$^{1}$\\
        &$E_{\phi}^{m} \uparrow$&.762$^{13}$&.760$^{14}$&.755$^{15}$&.780$^{12}$&.799$^{8}$&.784$^{10}$&.808$^{6}$&.705$^{17}$&.718$^{16}$&\textcolor{blue}{.819}$^{3}$&.810$^{5}$&.818$^{4}$&.798$^{9}$&.800$^{7}$&.800$^{7}$&.808$^{6}$&\textcolor{green}{.821}$^{2}$&.783$^{11}$&\textcolor{red}{.824}$^{1}$\\
        \hline
        \multirow{5}*{\textbf{OV}}
        &$maxF_{\beta} \uparrow$&.759$^{15}$&.756$^{16}$&.743$^{17}$&.797$^{12}$&.798$^{11}$&.768$^{14}$&.808$^{10}$&.696$^{18}$&.689$^{19}$&.818$^{7}$&.819$^{6}$&.816$^{8}$&.810$^{9}$&.796$^{13}$&.826$^{4}$&\textcolor{red}{.835}$^{1}$&\textcolor{green}{.830}$^{2}$&.823$^{5}$&\textcolor{blue}{.829}$^{3}$\\
        &$MAE \downarrow$&.173$^{13}$&.180$^{14}$&.184$^{15}$&.150$^{11}$&.136$^{8}$&.159$^{12}$&\textcolor{blue}{.125}$^{3}$&.216$^{16}$&.217$^{17}$&.129$^{6}$&.134$^{7}$&\textcolor{blue}{.125}$^{3}$&.146$^{9}$&.148$^{10}$&.126$^{4}$&\textcolor{green}{.119}$^{2}$&\textcolor{red}{.117}$^{1}$&.127$^{5}$&.126$^{4}$\\
        &$F_{\beta}^{w} \uparrow$&.637$^{13}$&.622$^{14}$&.616$^{15}$&.682$^{11}$&.701$^{9}$&.671$^{12}$&\textcolor{blue}{.733}$^{3}$&.527$^{17}$&.529$^{16}$&.723$^{5}$&.721$^{6}$&.724$^{4}$&.707$^{8}$&.697$^{10}$&.723$^{5}$&\textcolor{red}{.751}$^{1}$&\textcolor{green}{.738}$^{2}$&.720$^{7}$&.738$^{2}$\\
        &$S_{\alpha} \uparrow$&.721$^{15}$&.700$^{16}$&.688$^{17}$&.745$^{13}$&.751$^{10}$&.728$^{14}$&.762$^{7}$&.625$^{18}$&.611$^{19}$&.761$^{8}$&.748$^{11}$&.765$^{6}$&.752$^{9}$&.747$^{12}$&.774$^{4}$&\textcolor{green}{.779}$^{2}$&\textcolor{blue}{.775}$^{3}$&\textcolor{red}{.781}$^{1}$&.771$^{5}$\\
        &$E_{\phi}^{m} \uparrow$&.750$^{14}$&.737$^{15}$&.736$^{16}$&.778$^{13}$&.806$^{8}$&.789$^{12}$&\textcolor{green}{.828}$^{2}$&.663$^{18}$&.664$^{17}$&.816$^{4}$&.803$^{9}$&.809$^{6}$&.802$^{10}$&.795$^{11}$&.807$^{7}$&\textcolor{red}{.835}$^{1}$&\textcolor{blue}{.822}$^{3}$&.809$^{6}$&.814$^{5}$\\
        \hline
        \multirow{5}*{\textbf{SC}}
        &$maxF_{\beta} \uparrow$&.737$^{12}$&.735$^{14}$&.707$^{17}$&.764$^{10}$&.783$^{8}$&.71$^{16}$&.736$^{13}$&.697$^{18}$&.718$^{15}$&.780$^{9}$&.786$^{7}$&.793$^{4}$&.783$^{8}$&.790$^{5}$&\textcolor{blue}{.795}$^{3}$&.788$^{6}$&\textcolor{green}{.798}$^{2}$&.755$^{11}$&\textcolor{red}{.816}$^{1}$\\
        &$MAE \downarrow$&.098$^{12}$&.098$^{12}$&.101$^{14}$&.090$^{10}$&.081$^{7}$&.100$^{13}$&.087$^{9}$&.114$^{16}$&.110$^{15}$&\textcolor{blue}{.076}$^{3}$&.080$^{6}$&\textcolor{blue}{.076}$^{3}$&.083$^{8}$&\textcolor{green}{.075}$^{2}$&.078$^{5}$&.078$^{5}$&.077$^{4}$&.094$^{11}$&\textcolor{red}{.072}$^{1}$\\
        &$F_{\beta}^{w} \uparrow$&.608$^{15}$&.599$^{16}$&.593$^{18}$&.638$^{12}$&.677$^{10}$&.611$^{14}$&.669$^{11}$&.550$^{19}$&.594$^{17}$&.696$^{6}$&\textcolor{blue}{.708}$^{3}$&.701$^{5}$&.678$^{9}$&.695$^{7}$&.691$^{8}$&.705$^{4}$&\textcolor{green}{.711}$^{2}$&.626$^{13}$&\textcolor{red}{.749}$^{1}$\\
        &$S_{\alpha} \uparrow$&.768$^{10}$&.761$^{11}$&.745$^{13}$&.783$^{8}$&.799$^{5}$&.756$^{12}$&.772$^{9}$&.715$^{15}$&.724$^{14}$&\textcolor{blue}{.808}$^{3}$&.793$^{6}$&.807$^{4}$&.793$^{6}$&.807$^{4}$&\textcolor{green}{.809}$^{2}$&.807$^{4}$&\textcolor{blue}{.808}$^{3}$&.784$^{7}$&\textcolor{red}{.826}$^{1}$\\
        &$E_{\phi}^{m} \uparrow$&.793$^{14}$&.798$^{13}$&.787$^{16}$&.813$^{11}$&.840$^{9}$&.805$^{12}$&.837$^{10}$&.764$^{17}$&.791$^{15}$&.853$^{5}$&\textcolor{blue}{.858}$^{3}$&.848$^{7}$&.843$^{8}$&.856$^{4}$&.843$^{8}$&.850$^{6}$&\textcolor{green}{.859}$^{2}$&.798$^{13}$&\textcolor{red}{.869}$^{1}$\\
        \hline
        \multirow{5}*{\textbf{SO}}
        &$maxF_{\beta} \uparrow$&.664$^{14}$&.670$^{13}$&.663$^{15}$&.689$^{10}$&.685$^{11}$&.653$^{16}$&.683$^{12}$&.631$^{17}$&.653$^{16}$&.705$^{9}$&.712$^{8}$&.715$^{7}$&\textcolor{blue}{.735}$^{3}$&\textcolor{green}{.740}$^{2}$&.729$^{4}$&.719$^{6}$&.727$^{5}$&.705$^{9}$&\textcolor{red}{.753}$^{1}$\\
        &$MAE \downarrow$&.119$^{16}$&.109$^{11}$&.115$^{13}$&.099$^{10}$&.096$^{8}$&.116$^{14}$&.092$^{6}$&.118$^{15}$&.113$^{12}$&.089$^{4}$&.091$^{5}$&\textcolor{green}{.084}$^{2}$&.098$^{9}$&\textcolor{blue}{.087}$^{3}$&\textcolor{red}{.082}$^{1}$&.091$^{5}$&\textcolor{red}{.082}$^{1}$&.095$^{7}$&\textcolor{green}{.084}$^{2}$\\
        &$F_{\beta}^{w} \uparrow$&.523$^{17}$&.524$^{16}$&.526$^{15}$&.561$^{13}$&.567$^{11}$&.531$^{14}$&.602$^{8}$&.487$^{19}$&.518$^{18}$&.596$^{9}$&.623$^{4}$&.613$^{7}$&.594$^{10}$&\textcolor{green}{.626}$^{2}$&.614$^{6}$&.619$^{5}$&\textcolor{blue}{.624}$^{3}$&.565$^{12}$&\textcolor{red}{.656}$^{1}$\\
        &$S_{\alpha} \uparrow$&.718$^{14}$&.713$^{15}$&.703$^{17}$&.737$^{11}$&.732$^{13}$&.707$^{16}$&.736$^{12}$&.682$^{18}$&.682$^{18}$&.746$^{9}$&.745$^{10}$&.756$^{5}$&.749$^{7}$&\textcolor{green}{.768}$^{2}$&\textcolor{blue}{.767}$^{3}$&.755$^{6}$&.759$^{4}$&.748$^{8}$&\textcolor{red}{.774}$^{1}$\\
        &$E_{\phi}^{m} \uparrow$&.744$^{17}$&.755$^{14}$&.747$^{16}$&.769$^{11}$&.779$^{10}$&.751$^{15}$&.802$^{5}$&.731$^{18}$&.758$^{13}$&.791$^{8}$&.804$^{4}$&\textcolor{blue}{.806}$^{3}$&.784$^{9}$&\textcolor{red}{.814}$^{1}$&.796$^{7}$&.801$^{6}$&\textcolor{blue}{.806}$^{3}$&.765$^{12}$&\textcolor{green}{.813}$^{2}$\\
        \hline
        \multirow{5}*{\textbf{Avg.}}
        &$maxF_{\beta} \uparrow$&.728$^{12}$&.724$^{13}$&.714$^{15}$&.751$^{9}$&.749$^{10}$&.717$^{14}$&.745$^{11}$&.685$^{17}$&.692$^{16}$&.769$^{7}$&.773$^{6}$&.776$^{5}$&.778$^{4}$&.778$^{4}$&\textcolor{green}{.786}$^{2}$&\textcolor{blue}{.779}$^{3}$&\textcolor{green}{.786}$^{2}$&.766$^{8}$&\textcolor{red}{.799}$^{1}$\\
        &$MAE \downarrow$&.134$^{15}$&.133$^{14}$&.137$^{16}$&.118$^{12}$&.112$^{10}$&.128$^{13}$&.105$^{7}$&.152$^{17}$&.152$^{17}$&.103$^{5}$&.104$^{6}$&\textcolor{blue}{.099}$^{3}$&.115$^{11}$&.109$^{9}$&\textcolor{green}{.098}$^{2}$&.102$^{4}$&\textcolor{red}{.094}$^{1}$&.108$^{8}$&\textcolor{green}{.098}$^{2}$\\
        &$F_{\beta}^{w} \uparrow$&.598$^{15}$&.588$^{16}$&.586$^{17}$&.631$^{13}$&.641$^{12}$&.609$^{14}$&.670$^{8}$&.533$^{19}$&.550$^{18}$&.668$^{9}$&.685$^{4}$&.679$^{6}$&.658$^{10}$&.671$^{7}$&.680$^{5}$&\textcolor{blue}{.689}$^{3}$&\textcolor{green}{.692}$^{2}$&.642$^{11}$&\textcolor{red}{.712}$^{1}$\\
        &$S_{\alpha} \uparrow$&.738$^{12}$&.724$^{14}$&.713$^{15}$&.754$^{11}$&.754$^{11}$&.732$^{13}$&.757$^{10}$&.676$^{16}$&.669$^{17}$&.767$^{8}$&.762$^{9}$&.774$^{5}$&.762$^{9}$&.770$^{7}$&\textcolor{green}{.785}$^{2}$&.777$^{4}$&\textcolor{blue}{.780}$^{3}$&.772$^{6}$&\textcolor{red}{.788}$^{1}$\\
        &$E_{\phi}^{m} \uparrow$&.763$^{13}$&.761$^{14}$&.757$^{15}$&.785$^{11}$&.797$^{9}$&.778$^{12}$&.818$^{4}$&.721$^{17}$&.737$^{16}$&.812$^{7}$&.816$^{5}$&.818$^{4}$&.800$^{8}$&.813$^{6}$&.812$^{7}$&\textcolor{blue}{.820}$^{3}$&\textcolor{green}{.824}$^{2}$&.789$^{10}$&\textcolor{red}{.825}$^{1}$\\
        \hline
    \end{tabular}}
\end{table*}
\begin{figure*}[!t]
    \centering
    \includegraphics[width=\textwidth]{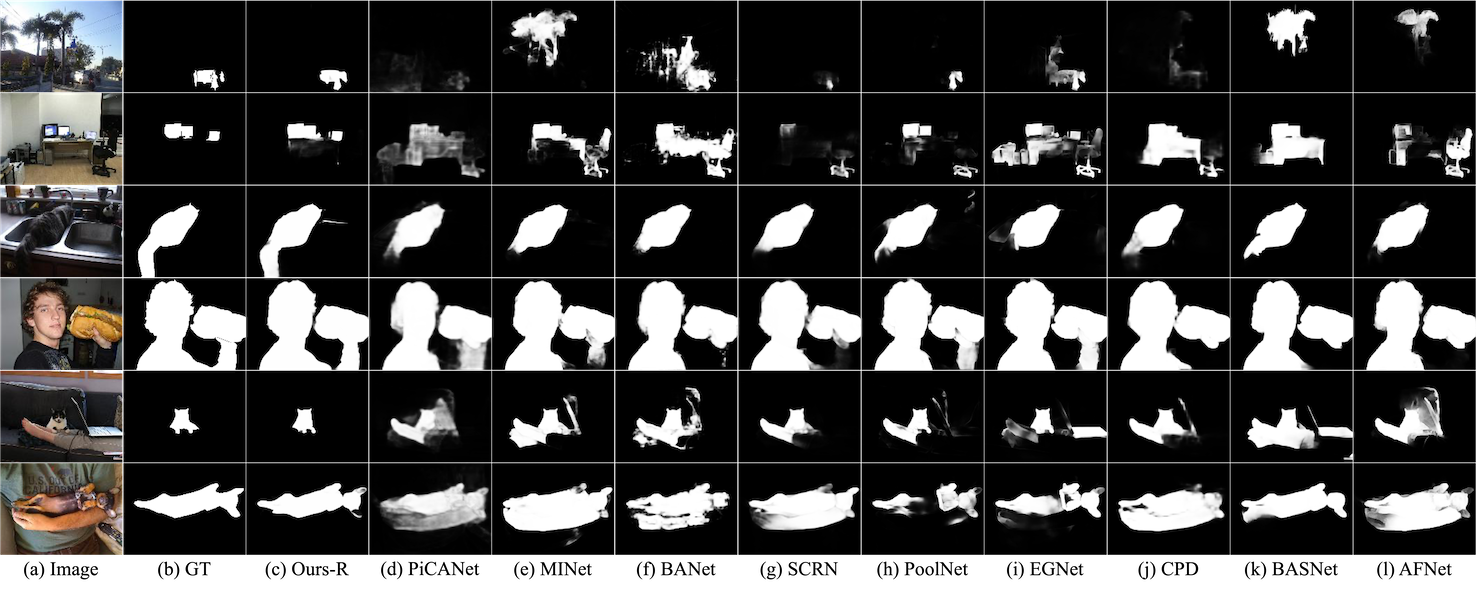}
    \caption{Low-resolution attribute-based qualitative comparison of the proposed method with nine other SOTA methods: (a) Image, (b) GT, (c) Ours-R, (d) PiCANet, (e) MINet, (f) BANet, (g) SCRN, (h) PoolNet, (i) EGNet, (j) CPD, (k) BASNet, (l) AFNet.}
    \label{fig:qualitative_soc}
\end{figure*}
\begin{figure*}[!t]
    \centering
    \includegraphics[width=\textwidth]{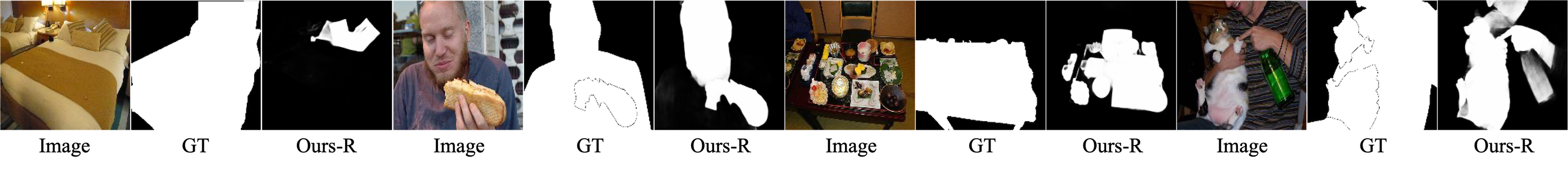}
    \caption{Failure cases of attribute-based analysis.}
    \label{fig:failurecase_SOC}
\end{figure*}
In comparing the ground truths (GTs) and Ours-Rs of the $1_{st}$ row of Fig. \ref{fig:failurecase}, we observe that our predicted saliency maps segment some objects in addition to the salient object in the GTs. However, these objects are crucial for providing contextual information, and we believe they possess similar saliency to the salient objects in the GTs. \textbf{In the process of dataset annotation, the photographer's intention must be considered}. For instance, the first example depicts a nail embedded in a tree trunk. In practical applications, segmenting only an overhead nail would destroy the image's original semantic information. The third image shows a child playing on a slide in a park, with the slide being crucial in reserving the meaning of the image, while the park is relatively unimportant and should be considered as the background. One might ask, what if I only want to keep the portrait in the image for replacing the background in practical application? We call this task as portrait matting \cite{shen2016deep} and it has corresponding datasets for the demand. For salient object detection (SOD) task, the objective is to segment the most salient object in the image, or in other words, the object that attracts your attention the most when you first look at the image. In the $2_{nd}$ row of Fig. \ref{fig:failurecase}, the salient objects in the GTs are completely opposite to the segmented objects in our predicted saliency maps. Our segmented objects are larger and have more distinct colors because larger and brighter objects tend to be more attention-grabbing. Moreover, we observe that in many datasets, for images that have both person and prominent landscapes, annotators tend to annotate only the person and consider the landscapes as background, even though these landscapes are what the photographer aims to highlight.

\begin{table*}[!t]
    \centering
    \scriptsize
    \caption{Data analysis of five high-resolution datasets. \textcolor{red}{Red}, \textcolor{green}{Green} and \textcolor{blue}{Blue} indicate the best, second best and third best.}\label{tab:hrdatasets_analysis}
    \renewcommand{\arraystretch}{1.0}
    \setlength\tabcolsep{1.0pt}
    \resizebox{\textwidth}{!}{
    \begin{tabular}{r|r|rrr|rrr}
        \hline
        \multirow{2}*{Dataset}&Number&\multicolumn{3}{c}{Image Dimension}\vline&\multicolumn{3}{c}{Object Complexity}\\
        \cline{2-8}
        &$I_{num}$&$H\pm \sigma_{H}$&$W\pm \sigma_{W}$&$D\pm \sigma_{D}$&$IPQ\pm \sigma_{IPQ}$&$C_{num}\pm \sigma_{C}$&$P_{num}\pm \sigma_{P}$\\
        \hline
        DIS5K \cite{qin2022highly}&5470&\textcolor{green}{2513.37}$\pm$ \textcolor{red}{1053.40}&\textcolor{green}{3111.44}$\pm$ \textcolor{green}{1359.51}&\textcolor{green}{4041.93}$\pm$ \textcolor{green}{1618.26}&\textcolor{red}{107.60}$\pm$ \textcolor{red}{320.69}&\textcolor{red}{106.84}$\pm$ \textcolor{red}{436.88}&\textcolor{red}{1427.82}$\pm$ \textcolor{red}{3326.72}\\

        ThinObject5K \cite{liew2021deep}&5748&1185.59$\pm$ \textcolor{blue}{909.53}&1325.06$\pm$ \textcolor{blue}{958.43}&1823.03$\pm$ \textcolor{blue}{1258.49}&\textcolor{green}{26.53}$\pm$ \textcolor{green}{119.98}&\textcolor{green}{33.06}$\pm$ \textcolor{green}{216.07}&\textcolor{blue}{519.14}$\pm$ \textcolor{green}{1298.54}\\

        UHRSD \cite{xie2022pyramid}&5920&\textcolor{blue}{1476.33}$\pm$ 272.34&1947.33$\pm$ 235.00&2469.12$\pm$ 67.11&\textcolor{blue}{9.62}$\pm$ \textcolor{blue}{30.71}&\textcolor{blue}{21.00}$\pm$ \textcolor{blue}{94.26}&\textcolor{green}{449.07}$\pm$ \textcolor{blue}{650.21}\\

        HRSOD \cite{zeng2019towards}&2010&\textcolor{red}{2713.12}$\pm$ \textcolor{green}{1041.70}&\textcolor{red}{3411.81}$\pm$ \textcolor{red}{1407.56}&\textcolor{red}{4405.40}$\pm$ \textcolor{red}{1631.03}&5.85$\pm$ 12.60&6.33$\pm$ 16.65&319.32$\pm$ 264.20\\

        DAVIS-S \cite{perazzi2016benchmark}&92&1299.13$\pm$ 440.77&\textcolor{blue}{2309.57}$\pm$ 783.59&\textcolor{blue}{2649.87}$\pm$ 899.05&7.84$\pm$ 5.69&15.60$\pm$ 29.51&389.58$\pm$ 309.29\\
        
        \hline
    \end{tabular}}
\end{table*}

\begin{table*}[!t]
    \centering
    \scriptsize
    \caption{Comparison of our method and 8 SOTA methods on DIS-TE, ThinObject5K, UHRSD, HRSOD, and DAVIS-S in terms of $HCE_{\gamma}$ ($\downarrow$), $mBA$ ($\uparrow$), $MAE$ ($\downarrow$), $F_{\beta}^{w}$ ($\uparrow$), and $S_{\alpha}$ ($\uparrow$). \textcolor{red}{Red}, \textcolor{green}{Green} and \textcolor{blue}{Blue} indicate the best, second best, and third best performance. The superscript of each score is the corresponding ranking.}\label{tab:quantitative_hr}
    \renewcommand{\arraystretch}{1.0}
    \setlength\tabcolsep{1.0pt}
    \resizebox{\textwidth}{!}{
    \begin{tabular}{r|cccc|lllll|lllll|lllll|lllll|lllll}
        \hline
        \multirow{2}*{\textbf{Method}}&\multirow{2}*{\textbf{Backbone}}&\multirow{2}*{\tabincell{c}{\textbf{Size}\\(MB)}}&\multirow{2}*{\tabincell{c}{\textbf{Input}\\\textbf{Size}}}&\multirow{2}*{\textbf{FPS}}&\multicolumn{5}{c}{\textbf{DIS-TE}(2470)}\vline&\multicolumn{5}{c}{\textbf{ThinObject5K}(5748)}\vline&\multicolumn{5}{c}{\textbf{UHRSD}(5920)}\vline&\multicolumn{5}{c}{\textbf{HRSOD}(2010)}\vline&\multicolumn{5}{c}{\textbf{DAVIS-S}(92)}\\
        &&&&&$HCE_{\gamma}$&$mBA$&$MAE$&$F_{\beta}^{w}$&$S_{\alpha}$&$HCE_{\gamma}$&$mBA$&$MAE$&$F_{\beta}^{w}$&$S_{\alpha}$&$HCE_{\gamma}$&$mBA$&$MAE$&$F_{\beta}^{w}$&$S_{\alpha}$&$HCE_{\gamma}$&$mBA$&$MAE$&$F_{\beta}^{w}$&$S_{\alpha}$&$HCE_{\gamma}$&$mBA$&$MAE$&$F_{\beta}^{w}$&$S_{\alpha}$\\
        \hline
        \textbf{SCRN}$_{19}$ \cite{wu2019stacked}&R-50&101.4&1024$\times$1024&32&1344$^{8}$&.703$^{8}$&.076$^{8}$&.685$^{9}$&.818$^{5}$&250$^{8}$&.744$^{9}$&.099$^{7}$&.743$^{7}$&.818$^{7}$&209$^{7}$&.732$^{6}$&.062$^{4}$&.807$^{7}$&.873$^{4}$&247$^{7}$&.694$^{7}$&\textcolor{green}{.068}$^{2}$&.757$^{4}$&\textcolor{green}{.855}$^{2}$&207$^{8}$&.718$^{9}$&\textcolor{blue}{.036}$^{3}$&.764$^{6}$&\textcolor{green}{.885}$^{2}$\\
        \textbf{CPD-R}$_{19}$ \cite{wu2019cascaded}&R-50&192.0&1024$\times$1024&28&1430$^{9}$&.696$^{9}$&.075$^{7}$&.694$^{7}$&.817$^{6}$&253$^{9}$&.752$^{8}$&.090$^{5}$&.761$^{5}$&.826$^{4}$&218$^{8}$&.728$^{7}$&.062$^{4}$&.814$^{4}$&.872$^{5}$&262$^{8}$&.694$^{7}$&.074$^{4}$&.746$^{5}$&.839$^{4}$&206$^{7}$&.731$^{8}$&\textcolor{blue}{.036}$^{3}$&.765$^{5}$&.880$^{4}$\\
        \textbf{F$^{3}$Net}$_{20}$ \cite{wei2020f3net}&R-50&102.5&1024$\times$1024&58&\textcolor{blue}{1115}$^{3}$&\textcolor{green}{.758}$^{2}$&\textcolor{blue}{.069}$^{3}$&.729$^{4}$&\textcolor{blue}{.821}$^{3}$&\textcolor{green}{188}$^{2}$&\textcolor{red}{.800}$^{1}$&.103$^{8}$&.741$^{8}$&.801$^{8}$&\textcolor{green}{173}$^{2}$&\textcolor{green}{.764}$^{2}$&.067$^{7}$&.804$^{8}$&.855$^{9}$&\textcolor{green}{208}$^{2}$&\textcolor{green}{.725}$^{2}$&.077$^{5}$&.743$^{6}$&.826$^{5}$&\textcolor{green}{160}$^{2}$&\textcolor{green}{.759}$^{2}$&.040$^{7}$&.738$^{9}$&.852$^{9}$\\
        \textbf{GCPANet}$_{20}$ \cite{chen2020global}&R-50&268.6&1024$\times$1024&60&1235$^{5}$&.727$^{6}$&.076$^{8}$&.692$^{8}$&.816$^{7}$&212$^{5}$&.781$^{4}$&.093$^{6}$&.757$^{6}$&.825$^{5}$&192$^{5}$&.748$^{5}$&.066$^{6}$&.799$^{9}$&.863$^{6}$&231$^{5}$&.704$^{6}$&.084$^{7}$&.711$^{8}$&.818$^{7}$&181$^{5}$&.746$^{5}$&.039$^{6}$&.742$^{8}$&.869$^{7}$\\
        \textbf{LDF}$_{20}$ \cite{wei2020label}&R-50&100.9&1024$\times$1024&58&1159$^{4}$&\textcolor{blue}{.751}$^{3}$&.070$^{4}$&\textcolor{blue}{.730}$^{3}$&.820$^{4}$&\textcolor{blue}{200}$^{3}$&\textcolor{blue}{.795}$^{3}$&.106$^{9}$&.736$^{9}$&.799$^{9}$&180$^{4}$&\textcolor{blue}{.761}$^{3}$&.064$^{5}$&.810$^{5}$&.860$^{7}$&218$^{4}$&.722$^{4}$&\textcolor{blue}{.072}$^{3}$&\textcolor{blue}{.761}$^{3}$&.839$^{4}$&173$^{4}$&\textcolor{blue}{.755}$^{3}$&.038$^{5}$&.747$^{7}$&.860$^{8}$\\
        \textbf{ICON-R}$_{21}$ \cite{zhuge2022salient}&R-50&132.8&1024$\times$1024&45&1337$^{7}$&.711$^{7}$&.072$^{6}$&.715$^{6}$&.809$^{8}$&249$^{7}$&.762$^{7}$&.089$^{4}$&.770$^{4}$&.822$^{6}$&221$^{9}$&.723$^{8}$&.069$^{8}$&.808$^{6}$&.856$^{8}$&269$^{9}$&.691$^{8}$&.080$^{6}$&.741$^{7}$&.824$^{6}$&206$^{7}$&.737$^{7}$&\textcolor{green}{.035}$^{2}$&.767$^{4}$&.870$^{6}$\\
        \textbf{PGNet}$_{22}$ \cite{xie2022pyramid}&R-18+S-B&279.2&1024$\times$1024&31&1285$^{6}$&.728$^{5}$&\textcolor{red}{.056}$^{1}$&\textcolor{red}{.769}$^{1}$&\textcolor{red}{.851}$^{1}$&226$^{6}$&.779$^{5}$&\textcolor{red}{.072}$^{1}$&\textcolor{red}{.806}$^{1}$&\textcolor{red}{.859}$^{1}$&196$^{6}$&.758$^{4}$&\textcolor{red}{.041}$^{1}$&\textcolor{red}{.872}$^{1}$&\textcolor{red}{.907}$^{1}$&238$^{6}$&\textcolor{blue}{.723}$^{3}$&\textcolor{red}{.041}$^{1}$&\textcolor{red}{.844}$^{1}$&\textcolor{red}{.898}$^{1}$&192$^{6}$&.738$^{6}$&\textcolor{red}{.023}$^{1}$&\textcolor{red}{.848}$^{1}$&\textcolor{red}{.917}$^{1}$\\
        \textbf{IS-Net}$_{22}$ \cite{qin2022highly}&RSU&176.6&1024$\times$1024&39&\textcolor{green}{1035}$^{2}$&.741$^{4}$&.071$^{5}$&.724$^{5}$&.818$^{5}$&202$^{4}$&.776$^{6}$&\textcolor{blue}{.085}$^{3}$&\textcolor{blue}{.778}$^{3}$&\textcolor{blue}{.831}$^{3}$&\textcolor{blue}{176}$^{3}$&\textcolor{blue}{.761}$^{3}$&\textcolor{blue}{.058}$^{3}$&\textcolor{blue}{.833}$^{3}$&\textcolor{blue}{.879}$^{3}$&\textcolor{blue}{211}$^{3}$&.712$^{5}$&\textcolor{blue}{.072}$^{3}$&.757$^{4}$&.839$^{4}$&\textcolor{blue}{171}$^{3}$&.748$^{4}$&\textcolor{blue}{.036}$^{3}$&\textcolor{blue}{.788}$^{3}$&\textcolor{blue}{.884}$^{3}$\\
        \rowcolor{gray!20}\textbf{DC-Net-R (Ours-R)}&R-34&356.3&1024$\times$1024&55&\textcolor{red}{984}$^{1}$&\textcolor{red}{.765}$^{1}$&\textcolor{green}{.063}$^{2}$&\textcolor{green}{.759}$^{2}$&\textcolor{green}{.840}$^{2}$&\textcolor{red}{180}$^{1}$&\textcolor{green}{.796}$^{2}$&\textcolor{green}{.081}$^{2}$&\textcolor{green}{.786}$^{2}$&\textcolor{green}{.839}$^{2}$&\textcolor{red}{163}$^{1}$&\textcolor{red}{.775}$^{1}$&\textcolor{green}{.053}$^{2}$&\textcolor{green}{.844}$^{2}$&\textcolor{green}{.885}$^{2}$&\textcolor{red}{203}$^{1}$&\textcolor{red}{.728}$^{1}$&\textcolor{green}{.068}$^{2}$&\textcolor{green}{.767}$^{2}$&\textcolor{blue}{.846}$^{3}$&\textcolor{red}{159}$^{1}$&\textcolor{red}{.763}$^{1}$&.037$^{4}$&\textcolor{green}{.789}$^{2}$&.879$^{5}$\\
        \hline
    \end{tabular}}
\end{table*}

\begin{figure*}[!t]
    \centering
    \includegraphics[width=\textwidth]{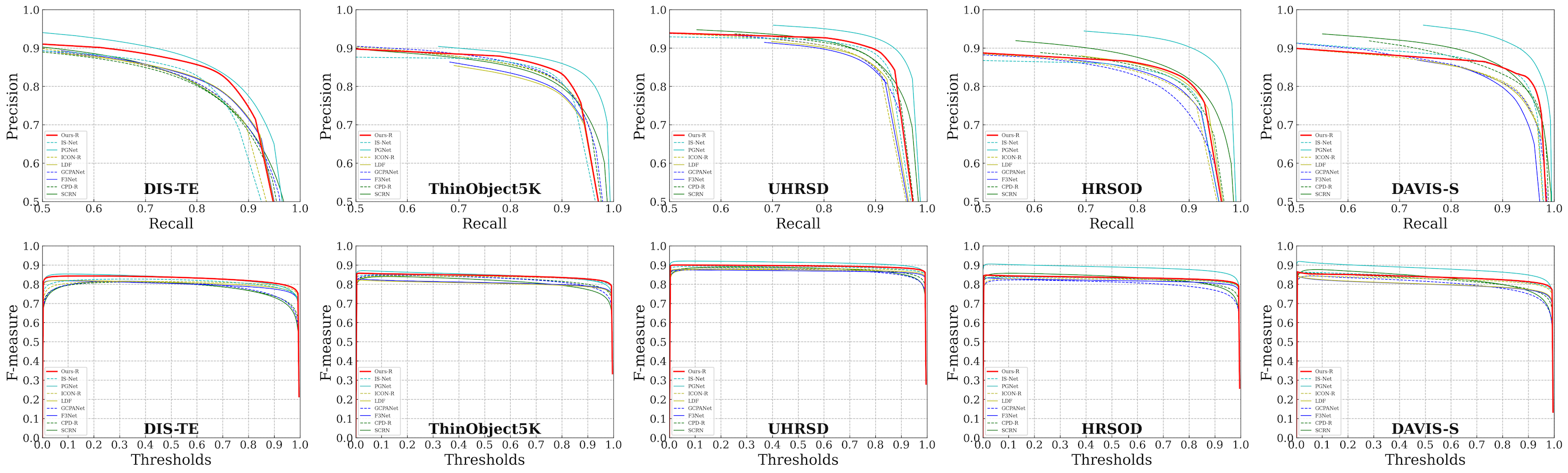}
    \caption{First row: Precision-Recall Curves comparison on five high-resolution saliency benchmark datasets. Second row: F-measure Curves comparison on five high-resolution saliency benchmark datasets.}
    \label{fig:pr_fm_hr}
\end{figure*}

\begin{figure*}[!h]
    \centering
    \includegraphics[width=\textwidth]{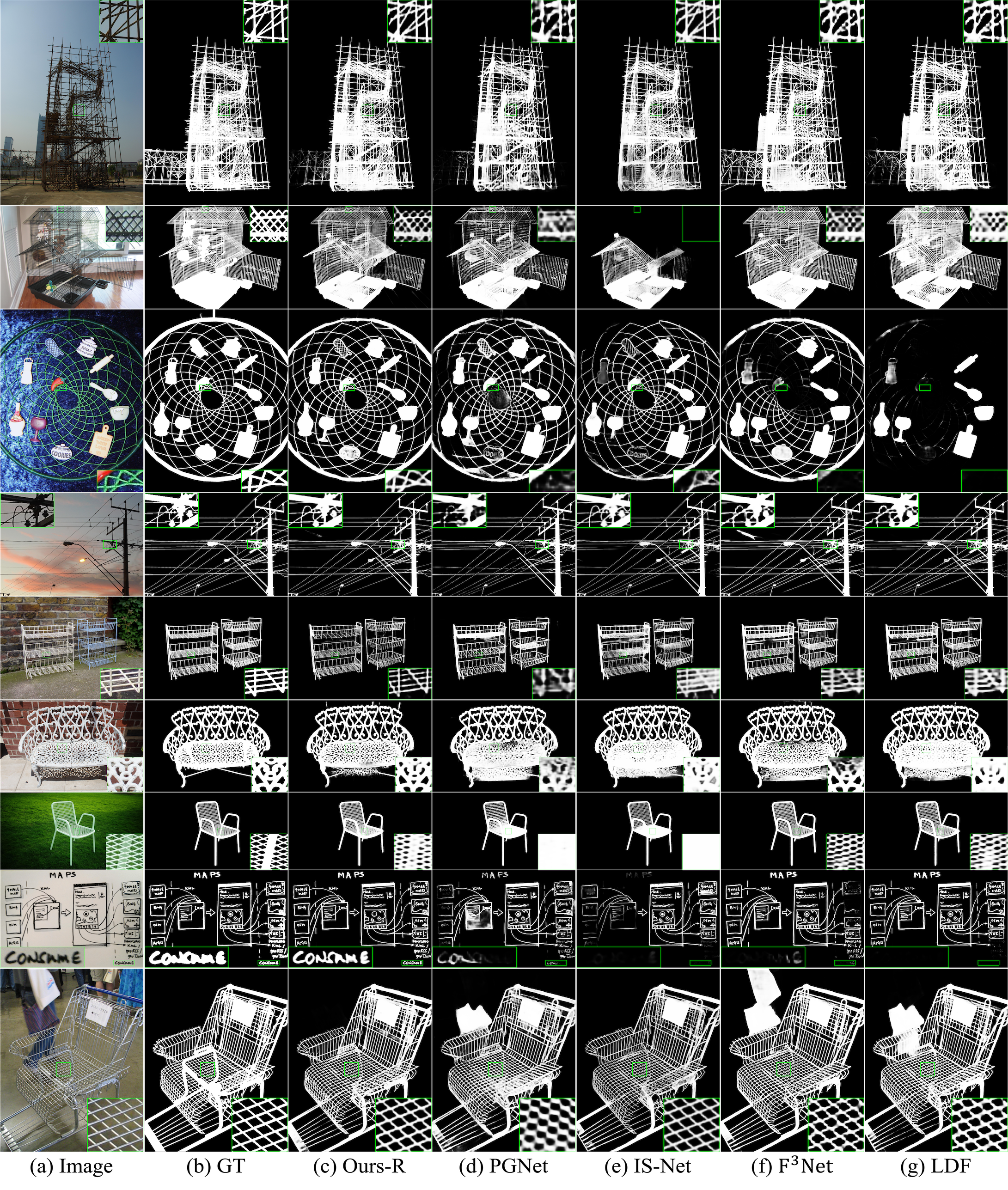}
    \caption{High-resolution qualitative comparison of the proposed method with four other SOTA methods: (a) Image, (b) GT, (c) Ours-R, (d) PGNet, (e) IS-Net, (f) F$^{3}$Net, (g) LDF.}
    \label{fig:qualitative_hr}
\end{figure*}

\subsubsubsection{Attribute-Based Analysis}
In addition to the previous 5 most frequently used saliency detection datasets, we also evaluate our DC-Net on another challenging SOC test dataset \cite{fan2022salient}. The SOC dataset divides images into the following nine groups according to nine different attributes: AC (Appearance Change), BO (Big Object), CL (Clutter), HO (Heterogeneous Object), MB (Motion Blur), OC (Occlusion), OV (Out-of-View), SC (Shape Complexity), and SO (Small Object).

We compare our DC-Net with 18 state-of-the-art methods, including Amulet \cite{zhang2017amulet}, DSS \cite{hou2017deeply}, NLDF \cite{luo2017non}, SRM \cite{wang2017stagewise}, BMPM \cite{zhang2018bi}, C2SNet \cite{li2018contour}, DGRL \cite{wang2018detect}, R$^{3}$Net \cite{deng2018r3net}, RANet \cite{chen2018reverse}, AFNet \cite{feng2019attentive}, BASNet \cite{qin2019basnet}, CPD \cite{wu2019cascaded}, EGNet \cite{zhao2019egnet}, PoolNet \cite{liu2019simple}, SCRN \cite{wu2019stacked}, BANet \cite{su2019selectivity}, MINet \cite{pang2020multi} and PiCANet \cite{liu2020picanet} in terms of attribute-based performance.

\textbf{Quantitative Comparison:}
Table \ref{tab:soc_quantitative} compares five evaluation metrics including $maxF_{\beta}$, $MAE$, $F_{\beta}^{w}$, $S_{\alpha}$ and $E_{\phi}^{m}$ of our proposed method with others. As we can see, our DC-Net achieves the state-of-the-art performance on attributes AC, CL, OC, SC, SO and their average in terms of almost all of the five evaluation metrics, and competitive performance on HO, MB and OV. On BO attribute, DC-Net performs relatively unremarkable and the cause of it is discussed in the failure cases part below. We calculate the average of nine attributes by $Avg.=\frac{\sum_{A}^{a}V_{a}\times N_{a}}{\sum_{A}^{a}N_{a}}$, where $A$ is the total number of attributes, $V_{a}$ is the $a_{th}$ metric value, and $N_{a}$ is the data amount of $a_{th}$ attribute. 

\textbf{Qualitative Comparison:}
Fig. \ref{fig:qualitative_soc} shows the sample results of our method and other nine best-performing methods in Table \ref{tab:soc_quantitative}, which intuitively demonstrates the promising performance of our method in three scenarios different from those mentioned in dataset-based analysis.

The salient objects depicted in the $1_{st}$ and $2_{nd}$ rows of Fig. \ref{fig:qualitative_soc} possess relatively modest saliency scores when contrasted with other images, but still maintain higher saliency compared to other objects within the same image. This leads to a challenging task for models to accurately detect them. Our method is capable of accurately localizing such objects. The $3_{rd}$ and $4_{th}$ rows exhibit results for salient objects with low-contrast, such as the tail of the cat in the third row and the arm in the fourth row. Our DC-Net-R demonstrates robustness in accurately segmenting these objects from the background. In the $5_{th}$ and $6_{th}$ rows, salient objects are occluded by surrounding confusing objects. By discerning the photographer's intent, it is apparent that the non-salient objects are not intended to draw attention in the image. Our method demonstrates accurate discrimination between salient and non-salient objects in such scenarios.

\textbf{Failure Cases:}
In the dataset-based analysis, we show that DC-Net-R has a good ability to segment large single salient objects, while the performance of DC-Net on the BO attribute is relatively unremarkable. We find that the BO test dataset contains many images which have both large and small salient objects in different categories, such as people holding food and different kinds of food on the table shown in Fig. \ref{fig:failurecase_SOC}. Our findings suggest that our method is better suited for segmenting salient objects of the same category, rather than handling scenarios with multiple salient objects belonging to different categories.

\begin{figure}[!t]
    \centering
    \includegraphics[scale=0.125]{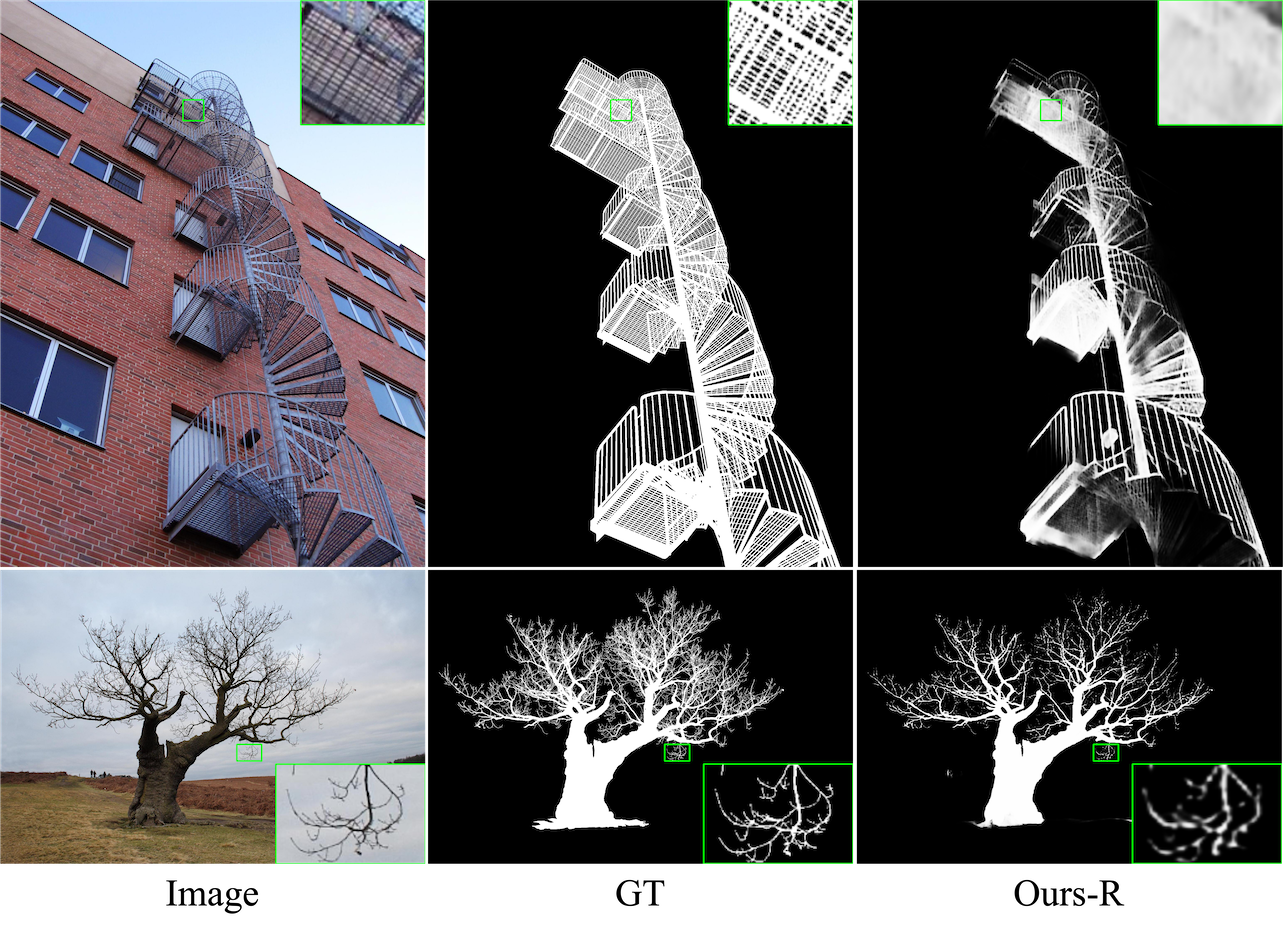}
    \caption{Failure cases of high-resolution images.}
    \label{fig:failurecase_hr}
\end{figure}
\subsection{Experiments on High-Resolution Saliency Detection Datasets}
As the results of the methods proposed by researchers on low-resolution datasets gradually become saturated, the development of high-resolution and high-quality (HH) segmentation has become an inevitable trend, especially for the meticulous fields of medical, aviation, and military. We suggest to use the following five datasets as training and evaluation datasets for HH methods: \textbf{DIS5K} \cite{qin2022highly}, \textbf{ThinObject5K} \cite{liew2021deep}, \textbf{UHRSD} \cite{xie2022pyramid}, \textbf{HRSOD} \cite{zeng2019towards} and \textbf{DAVIS-S} \cite{perazzi2016benchmark}. These datasets are all made for HH, Table \ref{tab:hrdatasets_analysis} shows their data analysis, which is calculated following \cite{qin2022highly}. $(H,W,D)$ and $(\sigma_{H},\sigma_{W},\sigma_{D})$ represent the mean of the image height, width, and diagonal length and their standard deviations respectively. The object complexity of datasets is evaluated by three metrics including the \textit{isoperimetric inequality quotient} ($IPQ \uparrow$) \cite{osserman1978isoperimetric,watson2011perimetric,yang2020meticulous}, the \textit{number of object contours} ($C_{num} \uparrow$) and the \textit{number of dominant points} ($P_{num} \uparrow$).

\subsubsection{Datasets}
\textbf{Training dataset:} \textbf{DIS5K} \cite{qin2022highly} can be seperated as a training dataset \textbf{DIS-TR}, a validation dataset \textbf{DIS-VD} and four test datasets \textbf{DIS-TE1}, \textbf{DIS-TE2}, \textbf{DIS-TE3} and \textbf{DIS-TE4}. We choose \textbf{DIS-TR} as our training dataset (3000 images) because its object complexity is much higher than other datasets. We believe that when the model can accurately segment complex objects, it becomes easier to segment simple objects.

\textbf{Evaluation datasets:} We evaluate our network on five benchmark datasets including: \textbf{DIS-TE} with 2470 images consisting of DIS-VD, DIS-TE1, DIS-TE2, DIS-TE3 and DIS-TE4, \textbf{ThinObject5K} \cite{liew2021deep} with 5748 images, \textbf{UHRSD} \cite{xie2022pyramid} with 5920 images, \textbf{HRSOD} \cite{zeng2019towards} with 2010 images, \textbf{DAVIS-S} \cite{perazzi2016benchmark} with 92 images.

\subsubsection{Comparison with State-of-the-arts}
We compare our DC-Net with 8 state-of-the-art methods including one \textbf{RSU} based model: \textbf{IS-Net}; one \textbf{ResNet-18} and \textbf{Swin-B} based model: \textbf{PGNet}; six \textbf{ResNet-50} based model: \textbf{SCRN}, \textbf{F$^{3}$Net}, \textbf{GCPANet}, \textbf{LDF}, \textbf{ICON-R}, \textbf{CPD-R}, we selected their better models based on \textbf{ResNet} or \textbf{VGG} for comparison. For a fair comparison, we run the official implementation of \textbf{IS-Net} which is trained on \textbf{DIS-TR} with pre-trained model parameters provided by the author to evaluate with the same evaluation code. Moreover, we re-train \textbf{PGNet}, \textbf{SCRN}, \textbf{F$^{3}$Net}, \textbf{GCPANet}, \textbf{LDF}, \textbf{ICON-R}, and \textbf{CPD-R} on \textbf{DIS-TR} based on their official implementation provided by the authors. We choose the above methods since their source codes have great reproducibility. Among them, \textbf{IS-Net} and \textbf{PGNet} are designed for high resolution, and others are designed for low resolution.

\textbf{Quantitative Comparison:}
Table \ref{tab:quantitative_hr} compares five evaluation metrics including $HCE_{\gamma}$, $mBA$, $MAE$, $F_{\beta}^{w}$, and $S_{\alpha}$ of our proposed method with others, where $HCE_{\gamma}$ and $mBA$ are designed for evaluating the detail quality of high-resolution saliency maps. As we can see, our DC-Net-R achieves state-of-the-art performance on almost all datasets in terms of $HCE_{\gamma}$ and $mBA$, and the second-best performance on DIS-TE, ThinObject5K, UHRSD, and HRSOD in terms of $MAE$, $F_{\beta}^{w}$, and $S_{\alpha}$. 
We find that PGNet obtain SOTA results on all datasets in terms of $MAE$, $F_{\beta}^{w}$, and $S_{\alpha}$ and unremarkable results on $HCE_{\gamma}$ and $mBA$, which indicate that Swin Transformer outperforms ResNet in detection but may not excel in capturing details.
The Fig. \ref{fig:pr_fm_hr} illustrates the precision-recall curves and F-measure curves which are consistent with the Table \ref{tab:quantitative_hr}. 

\textbf{Qualitative Comparison:}
Fig. \ref{fig:qualitative_hr} shows the sample results of our method and the other four best-performing methods in Table \ref{tab:quantitative_hr}, which intuitively demonstrates that our method can also achieve promising results on high-resolution datasets. Ours not only accurately detects salient objects but also produces smooth and high-confidence segmentation results for fine and dichotomous parts. In contrast, the segmentation results of PGNet, F$^{3}$Net, and LDF appear rough. Although the detail quality of IS-Net is competitive, the confidence level is slightly lower. Specifically, the $3_{rd}$, $8_{th}$, and $9_{th}$ rows display large objects that almost occupy the entire image, while other methods either miss some parts or segment out incorrect parts. In contrast, our method can accurately segment them, demonstrating that the large and compact receptive field provided by $ResASPP^{2}$ enables the model with the ability to recognize holistic semantics.

\textbf{Failure Cases:}
As shown in Fig. \ref{fig:failurecase_hr}, both the Image and GT are displayed at the original pixel size, whereas the saliency map is obtained by downsampling the original image to $1024\times 1024$ and then processing it through the model. As a result, a significant amount of precision and detail is lost, especially for extremely small parts. The spiral iron stair in $1_{st}$ row has densely staggered parts, resulting in a lot of holes of different sizes interspersed between the iron stairs. It is difficult for our method to segment such a dichotomous object with the input size of $1024\times 1024$. The branches in $2_{nd}$ row is a difficult case for highly accurate segmentation. It has the characteristics of irregular shape, uncertain direction, and meticulosity, which makes the confidence of predicted saliency maps low. Therefore, models that can handle higher-resolution input images to obtain detailed object structures, with acceptable memory usage, training and inference time costs on the mainstream GPUs are needed.

\section{Conclusion}
In this paper, we propose a novel salient object detection model DC-Net. Our DC-Net explicitly guides the model's training process by using the concept of Divide-and-Conquer, and then obtains larger and more compact effective receptive fields (ERF) and richer multi-scale information through our newly designed two-level Residual nested-ASPP (ResASPP$^{2}$) modules. Additionally, we hope that our parallel version of ResNet and Swin-Transformer can promote the research of multiple encoder models. Experimental results on six public low-resolution and five high-resolution salient object detection datasets demonstrate that our DC-Net achieves competitive performance against 21 and 8 state-of-the-art methods respectively. We also demonstrate through experiments that edge maps with different edge widths have a significant impact on the model's performance.

Although our model achieves competitive results compared to other state-of-the-art methods, the disadvantage of using multiple encoders leads to an increase in parameters. In the near future, we will explore different techniques such as distillation to address this issue. Furthermore, as mentioned above, how to find reasonable auxiliary map combinations for more encoders and how to enable the model to handle larger resolution input images in acceptable memory usage, training and inference time costs are also urgent issues to be addressed.

{\small
\bibliographystyle{ieee_fullname}

}

\end{document}